\documentclass{article}

\usepackage[dblblindworkshop,final]{neurips_2026}

\usepackage[utf8]{inputenc} 
\usepackage[T1]{fontenc}    
\usepackage{hyperref}       
\usepackage{url}            
\usepackage{booktabs}       
\usepackage{amsfonts}       
\usepackage{nicefrac}       
\usepackage{microtype}      
\usepackage{xcolor}         
\usepackage{amsmath}
\usepackage{algorithm}
\usepackage{algpseudocode}
\usepackage{graphicx}
\usepackage{float}
\usepackage{caption}
 \usepackage{multirow}

\newcommand{\method}{TRACE}

\title{TRACE: Adaptive Trust-Region Bayesian Optimization for Open-Ended Molecular Design}
\workshoptitle{AI for Accelerated Materials Design (AI4Mat)}

\author{%
  Madhav R. Muthyala \\
  UW-Madison \\
  \texttt{mmuthyala@wisc.edu} \\
  \And
  Mahsa Askari \\
  Texas Tech University \\
  \texttt{Mahsa.Askari@ttu.edu} \\
  \And
  Farshud Sorourifar \\
  The Ohio State University \\
  \texttt{sorourifar.1@osu.edu} \\
  \And
  Evan Zhou \\
  UW-Madison \\
  \texttt{ezhou37@wisc.edu} \\
  \And
  Joe Gauthier \\
  Texas Tech University \\
  \texttt{Joe.Gauthier@ttu.edu} \\
  \And
  Joel A. Paulson \\
  UW-Madison \\
  \texttt{joel.paulson@wisc.edu} \\
}

\begin{document}

\maketitle

\begin{abstract}
Optimizing molecules under small evaluation budgets requires deciding both where in chemical space to search and which candidate within that region to evaluate. We introduce Trust Region Adaptive Chemical Exploration (TRACE), a framework for open-ended molecular Bayesian optimization (BO), where candidates are generated on demand rather than selected from a fixed library. TRACE combines an objective-free graph-perturbation mechanism with a fixed, chemistry-aware Gaussian process (GP) surrogate using fragprints, a representation combining molecular fingerprints and fragment descriptors. Rather than learning the search geometry itself, TRACE makes the scale of chemical exploration explicit through a Tanimoto-similarity threshold controlling chemical extent and a cardinality budget controlling candidate-search breadth. A simple success/failure controller expands or contracts both quantities during optimization, while the GP and acquisition function select which admissible molecules to evaluate. Across six challenging molecular optimization benchmarks under matched 1,000-evaluation budgets, TRACE achieves the strongest overall sample efficiency as measured by the best molecule found as evaluations accumulate. Crossed ablations on two objectives further show that adaptive region control contributes beyond the proposal mechanism, while its benefit depends on the proposal distribution. 
We further apply TRACE to the stability of iron phthalocyanine (FePc), a non-precious molecular electrocatalyst whose degradation can involve loss of the Fe center. Using a first-principles density functional theory (DFT) workflow designed for accurate energy evaluation, TRACE increases the computed thermodynamic resistance to Fe dissolution from 0.42 to 2.59 eV over 80 designed evaluations.
\end{abstract}

\section{Introduction}
\label{sec:intro}

Molecular optimization is a fundamental challenge in chemistry, with applications ranging from drug discovery to catalyst design \citep{mpo, muthyala2026generative, casillo2025catalysis}. In the ``open-ended'' setting considered here, candidate molecules are generated on demand rather than selected from a fixed, pre-enumerated library. Optimization then involves two intertwined decisions, specifically, where in chemical space to search and which candidate within that region to evaluate next. The second has received substantial attention through surrogate models and acquisition functions.
The first is often encoded in the candidate-generation mechanism, representation-dependent neighborhoods, similarity criteria, or the number of candidates proposed per iteration.
Graph-based genetic algorithms traverse chemical space through prescribed mutation and crossover operations \citep{jensen2019graph}; STONED constructs local neighborhoods through discrete modifications of a string representation \citep{nigam2021beyond}. These choices often encode considered chemical judgment about what constitutes a sensible neighborhood. In many molecular optimization pipelines, these choices are specified before optimization and remain fixed throughout the run. They nonetheless determine the effective scale of exploration, and thus how a limited oracle budget is distributed across a combinatorially large space.

On the Practical Molecular Optimization (PMO) benchmark, most molecular design algorithms introduced as state of the art fail to outperform substantially simpler predecessors under a fixed allowance of $10^4$ oracle calls \citep{gao2022sample}. Even this budget is large relative to many discovery settings. A first-principles electronic structure calculation for a candidate catalyst can require hours of computation, while synthesis and experimental characterization can take days or longer. When evaluation dominates cost, the scale of exploration can strongly affect how the oracle budget is used. Searching too narrowly can concentrate evaluations on marginal variations of a local scaffold, while searching too broadly requires the optimizer to discriminate among candidates spanning a larger and more heterogeneous portion of chemical space. The appropriate scale may also vary with the task, the region currently under consideration, and the observations accumulated during optimization. Bayesian optimization (BO) is well suited to such settings because it uses a probabilistic surrogate and acquisition function to guide expensive evaluations \citep{paulson2025bayesian}. 
GAUCHE \citep{griffiths2023gauche} showed that Tanimoto-kernel Gaussian processes (GPs) using fragprints, which concatenate Morgan fingerprints with RDKit fragment descriptors, perform well in predictive accuracy, uncertainty calibration, and runtime on small- to mid-sized molecular datasets, and demonstrated their use in data-limited molecular BO.
A separate study \citep{tripp2024diagnosing} showed that after correcting an incorrect prior width, over-smoothing, and inadequate acquisition-function maximization, a basic GP-based BO setup attains the highest overall performance on PMO under their 10,000-call evaluation protocol. Their remedy for inadequate acquisition maximization uses a genetic algorithm to search molecular space more extensively, highlighting the role of candidate generation in the acquisition stage. Together, these results motivate retaining a lightweight molecular GP while making the scale of candidate search adaptive.

In (high-dimensional) continuous domains, TuRBO \citep{eriksson2019scalable} provides a useful template for adapting the scale of local BO from optimization progress. It restricts search to a hyperrectangle around the incumbent and expands/contracts that region after consecutive successes/failures. The analogous choice in molecular design is less straightforward because locality depends strongly on the representation. Categorical trust region methods use Hamming distance over fixed-length categorical or mixed representations \citep{wan2021think, papenmeier2023bounce}, which does not directly encode structural similarity for variable-size molecular graphs. Latent-space molecular BO instead defines local regions in learned embeddings \citep{tripp2020sample, maus2022local, lee2023cobo, chu2024invbo}. Recent methods have addressed important aspects of the mismatch between latent geometry and the underlying molecular or objective space, but the resulting notion of locality remains tied to a learned representation. 
Other methods explore chemical space through genetic search or learned molecular generators \citep{jensen2019graph, kim2024genetic, guo2024augmented}. TRACE instead makes chemical search scale explicit by adapting the similarity-defined region and candidate set passed to the acquisition function.


A fingerprint-similarity threshold, however, does not fully specify the finite candidate set over which BO operates. The threshold $\tau$ controls \emph{chemical extent}, or how dissimilar an admitted molecule may be from the current reference, whereas a cardinality budget $k$ controls \emph{candidate-search breadth}, or how many admissible molecules are passed to the acquisition function. Because the number of generated molecules satisfying a given threshold depends on the current reference and proposal distribution, the same value of $\tau$ can correspond to very different candidate set sizes. TRACE therefore couples $\tau$ and $k$ through a simple monotone map, giving a single adaptive state that expands or contracts both chemical extent and candidate-search breadth. This coupling is a practical design choice, and other mappings or separate controllers are possible. Fingerprint similarity is also only an imperfect proxy for property similarity, as illustrated by activity cliffs between structurally similar molecules with sharply different properties \citep{maggiora2006outliers}. Rather than fixing one nominal similarity scale for the entire run, TRACE adjusts the search scale in response to optimization progress.

\begin{figure*}[t]
    \centering
    \includegraphics[width=0.98\textwidth]{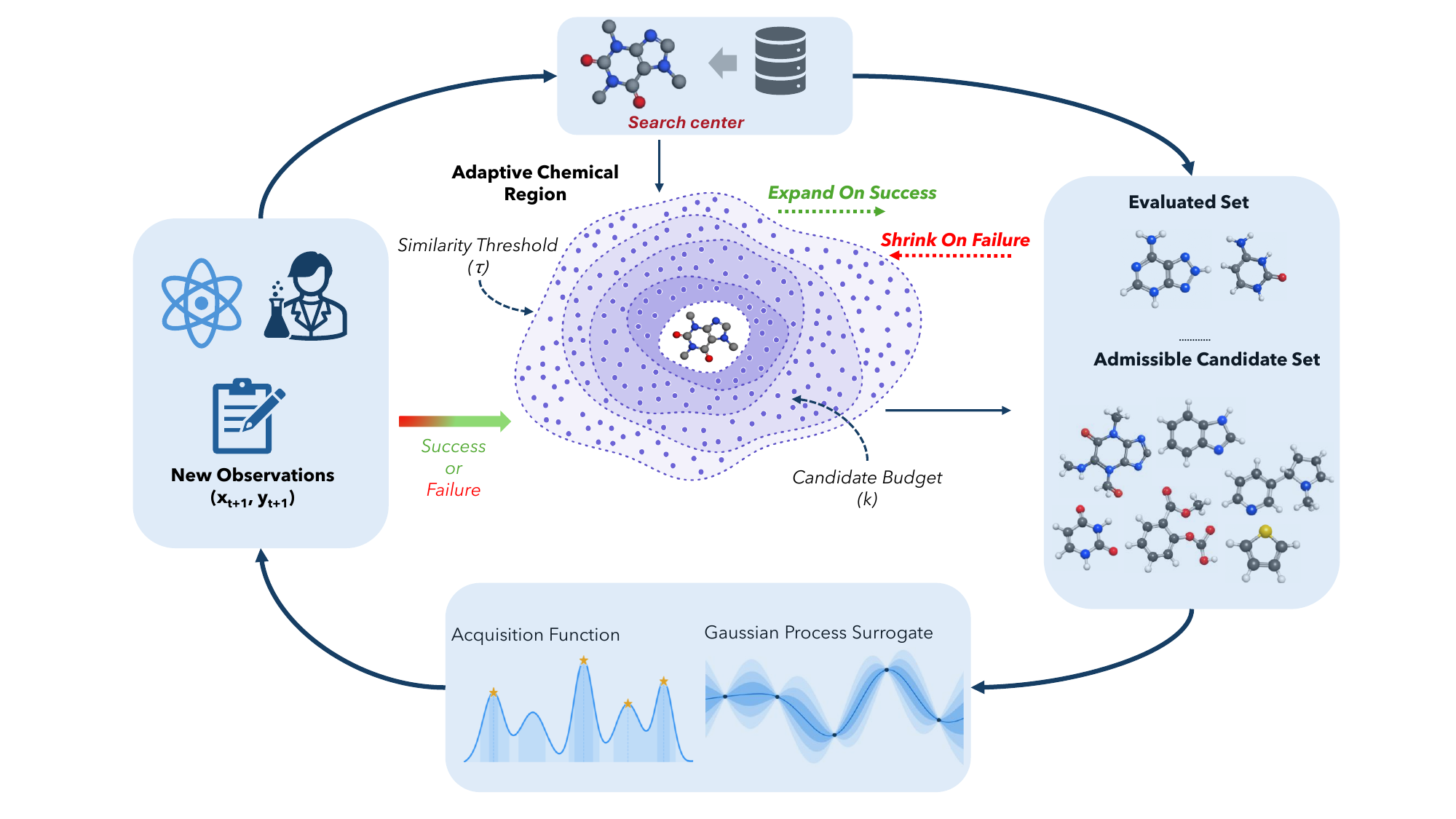}
    \caption{\textbf{Overview of the TRACE framework.}
    At iteration $t$, an objective-free proposal mechanism generates candidates around the current search center. The adaptive chemical region admits
    candidates according to a Tanimoto-similarity threshold $\tau_t$ and cardinality budget $k_t$, which are updated by a common success/failure controller. Predictions from a GP surrogate fitted to the evaluated set $\mathcal{D}_t$ are used by a batch acquisition function to select molecules from the admissible set $\mathcal{A}_t$ for oracle evaluation. The resulting observations update the evaluated set and the incumbent, and whether the batch meets the controller success criterion provides the signal used to expand or contract the region.}
    \label{fig:trace_framework}
\end{figure*}

We introduce \textbf{TRACE} (\textbf{T}rust \textbf{R}egion \textbf{A}daptive \textbf{C}hemical \textbf{E}xploration), a framework for open-ended molecular BO that makes the scale of candidate search explicit and adaptive. An objective-free graph-perturbation mechanism generates molecular proposals around the current search center. TRACE filters these proposals using a Tanimoto-similarity threshold $\tau$ that controls chemical extent and a cardinality budget $k$ that controls candidate-search breadth, with a common success/failure controller expanding or contracting both during optimization. A Tanimoto-kernel GP over fragprints \citep{griffiths2023gauche} and an acquisition function then select molecules from the admissible set for evaluation. Figure~\ref{fig:trace_framework} summarizes this division of roles between proposal generation, adaptive region control, and objective-aware candidate selection. Across six PMO objectives, TRACE shows strong sample efficiency under a matched low-budget protocol, and crossed ablations on two objectives show that the adaptive region contributes beyond the proposal mechanism while providing no corresponding benefit with Graph-GA proposals. In a closed-loop density functional theory (DFT) campaign on iron phthalocyanine (FePc) derivatives, TRACE improves the computed Fe-dissolution-resistance objective by 2.17 eV over 80 designed evaluations while moving substantially away from the initial molecule.
This expensive-oracle setting is directly relevant to AI-guided materials design, where efficiently allocating limited high-fidelity evaluations can help accelerate materials discovery.

\section{TRACE Design}
\label{sec:trace_design}

\subsection{Search center and admissible set}

Let $\mathcal{X}$ denote chemical space, $f:\mathcal{X}\rightarrow\mathbb{R}$ the oracle, and $\mathcal{D}_t$ the evaluated data available at the start of iteration $t$, consisting of molecules and their observed oracle values. We write $\mathcal{X}(\mathcal{D}_t)$ for the molecules in $\mathcal{D}_t$ and let $x_t^{\mathrm{best}}$ denote the \emph{global incumbent}, the best molecule evaluated so far. TRACE also maintains a \emph{search center} $c_t$ used to generate and filter proposals. In ordinary iterations $c_t=x_t^{\mathrm{best}}$; a restart may instead select another previously evaluated molecule as the search center without changing the global incumbent.

We measure structural similarity with the Tanimoto coefficient $T(x,x')$ computed on binary Morgan fingerprints with radius 2 and 2048 bits \citep{rogers2010extended}. A Tanimoto-similarity threshold $\tau_t$ defines the conceptual chemical region around the search center,
\begin{equation}
\mathcal{R}_t=
\left\{
x\in\mathcal{X}:T(x,c_t)\geq\tau_t
\right\}.
\label{eq:neighborhood}
\end{equation}
TRACE does not enumerate $\mathcal{R}_t$. Instead, the proposal mechanism supplies a finite, deduplicated set $\mathcal{P}_t$. After removing evaluated molecules and applying the similarity threshold, let
\begin{equation}
\begin{aligned}
\mathcal{S}_t
&=
\left\{
x\in\mathcal{P}_t\setminus\mathcal{X}(\mathcal{D}_t):
T(x,c_t)\geq\tau_t
\right\},\\
\mathcal{A}_t
&\subseteq\mathcal{S}_t,
\qquad
|\mathcal{A}_t|=\min(k_t,|\mathcal{S}_t|).
\end{aligned}
\label{eq:budget}
\end{equation}
When $|\mathcal{S}_t|>k_t$, the admissible set $\mathcal{A}_t$ contains the $k_t$ molecules in $\mathcal{S}_t$ with largest $T(x,c_t)$; otherwise $\mathcal{A}_t=\mathcal{S}_t$. Thus, $\tau_t$ controls \emph{chemical extent}, while the cardinality budget $k_t$ controls \emph{candidate-search breadth}. The admissible set determines where objective-aware selection operates: the GP posterior and a batch acquisition function $\alpha_t$ select a batch $B_t\subseteq\mathcal{A}_t$ with $|B_t|=q_t$, where $q_t$ is the number of oracle evaluations available in the current batch (normally the nominal batch size $q$, reduced only when fewer evaluations remain in the total budget).

\subsection{Adaptive region control}

TRACE uses $k_t\in[k_{\min},k_{\max}]$ as a single adaptive state. Let $\tau_{\mathrm{hi}}$ and $\tau_{\mathrm{lo}}$ denote the largest and smallest allowed similarity thresholds, respectively, with $0\leq\tau_{\mathrm{lo}}<\tau_{\mathrm{hi}}\leq1$, and let $k_0\in[k_{\min},k_{\max}]$ be the initial cardinality budget. TRACE couples $\tau$ to $k$ through
\begin{equation}
\tau(k)
=
\tau_{\mathrm{hi}}
-
\frac{k-k_{\min}}{k_{\max}-k_{\min}}
\left(\tau_{\mathrm{hi}}-\tau_{\mathrm{lo}}\right).
\label{eq:coupling}
\end{equation}
Hence, $\tau(k_{\min})=\tau_{\mathrm{hi}}$ and $\tau(k_{\max})=\tau_{\mathrm{lo}}$. Expanding the region increases $k$ and lowers $\tau$, increasing candidate-search breadth while allowing more chemically distant proposals. On the other hand, contracting the region decreases $k$ and raises $\tau$, concentrating search nearer the center. Eq.~\eqref{eq:coupling} is a deliberately simple monotone coupling that gives TRACE one adaptive state; other mappings or separate controllers are possible.

After each evaluated batch, TRACE updates $x_{t+1}^{\mathrm{best}}$ to the best molecule in $\mathcal{D}_{t+1}$, regardless of the size of the improvement. For a relative tolerance $\varepsilon\geq0$, the iteration is a controller success when
\[
f(x_{t+1}^{\mathrm{best}})
>
f(x_t^{\mathrm{best}})
+
\varepsilon
\max\!\left(1,\left|f(x_t^{\mathrm{best}})\right|\right),
\]
and is otherwise a failure. After $n_{\mathrm{succ}}$ consecutive successes, TRACE expands by setting $k\leftarrow\min(k_{\max},2k)$; after $n_{\mathrm{fail}}$ consecutive failures, it contracts by halving $k$. Here $n_{\mathrm{succ}}$ and $n_{\mathrm{fail}}$ are the corresponding consecutive-event thresholds. If another contraction would place $k$ below the feasible batch floor $\max(k_{\min},q_t)$, TRACE restarts instead, resetting $k$ to $k_0$ and choosing a new search center from evaluated molecules. A restart can also occur when the current proposal supply cannot fill a batch: TRACE first generates an additional relaxed, objective-free fallback pool and restarts only if the augmented admissible set remains underfilled. Algorithm~\ref{alg:trace} in Appendix~\ref{app:trace_outer} gives the complete state-transition logic and restart rule. Implementation details and code are available at \url{https://github.com/PaulsonLab/TRACE-AI4MAT}. 

\subsection{Objective-free proposal generation and BO selection}
\label{subsec:proposal}

The proposal mechanism generates structural variation without observing oracle values. It applies chemistry-aware atom, bond, fragment, and ring edits under constraints derived from the center used to generate each proposal pool. The proposer may adapt its internal behavior using structural and proposal-history information, including operator outcomes, locality, shell coverage, novelty, and molecular complexity. Because the search center is supplied by the outer optimization loop, proposal generation depends indirectly on optimization history; ``objective-free'' means that the proposal routine itself never observes $f(x)$ or uses objective values to rank molecules.

Each iteration combines a fresh strict local pool generated around $c_t$ with a broader relaxed reservoir retained from initialization or the most recent restart. If too few molecules survive admission, a temporary relaxed fallback pool augments the proposal set before TRACE considers restarting. Algorithm~\ref{alg:gen} in Appendix~\ref{app:proposal_algorithm} gives the proposal routine and the roles of these constraint profiles.

For objective-aware selection, TRACE fits a GP to all observations in $\mathcal{D}_t$ using fragprints, which concatenate Morgan fingerprints with RDKit fragment descriptors, and a Tanimoto kernel \citep{griffiths2023gauche}. This surrogate representation is distinct from the radius-2, 2048-bit Morgan fingerprints used to define $T$. A batch acquisition function $\alpha_t$ computed from the GP posterior is optimized over the finite admissible set $\mathcal{A}_t$ to select $B_t$. The experiment-specific acquisition choices are reported with the corresponding experimental protocols.


\section{Experiments}
\label{sec:experiments}

\subsection{Low-budget molecular optimization benchmarks}
\label{sec:low_budget_benchmarks}

We first test whether adaptive control of chemical search scale translates into improved sample efficiency on established molecular optimization objectives. We select six objectives from the Practical Molecular Optimization (PMO) benchmark \citep{gao2022sample}: QED, median molecule 2, osimertinib MPO, scaffold hopping, decoration hopping, and DRD2. We chose these objectives to cover different optimization tasks with limited redundancy: single-property drug-likeness (QED), multi-target similarity (median molecule 2), multi-property objectives (osimertinib MPO), structural hopping (scaffold and decoration hopping, which differ in what is held fixed), and a learned activity classifier (DRD2). Our aim is not to reproduce the full PMO leaderboard, which contains 23 objectives and uses a substantially larger oracle budget, but to ask whether a consistent low-budget signal emerges when the methods are compared under the same evaluation constraints.


For each objective and seed, all methods start from the same 20 evaluated molecules. They then receive 20 rounds of active optimization with batch size $q=49$, for a total budget of 1,000 unique-molecule oracle evaluations including initialization. We use five seeds and apply the same oracle and molecular constraint filter (SA $\leq 7$; SMILES length $\leq 100$) throughout. For \method{}, the Tanimoto-kernel GP described in Section~\ref{sec:trace_design} is refit as observations accumulate, and qLogEI \citep{ament2023unexpected} is
optimized over the current admissible set to select each batch. qLogEI is the batch form of log expected improvement, a numerically stabilized reformulation of expected improvement designed to preserve its improvement-seeking behavior while making the acquisition function easier to optimize.

We compare against methods that represent several distinct strategies for exploring chemical space. Graph-GA \citep{jensen2019graph} and SMILES-GA \citep{brown2019guacamol} provide strong evolutionary baselines that search directly through molecular graph and string modifications, respectively. STONED-SELFIES \citep{nigam2021beyond} instead explores local neighborhoods through discrete SELFIES edits, providing a useful comparison to explicit local molecular search without a probabilistic surrogate. REINVENT4 \citep{loeffler2024reinvent} represents a modern reinforcement learning (RL) approach in which a generative policy is adapted toward high-scoring molecules.
Two baselines are particularly close to the questions motivating \method{}. GP-BO + Graph-GA \citep{tripp2024diagnosing} combines molecular GP-based BO with a graph genetic algorithm for acquisition-function optimization, providing a strong surrogate-guided comparison without \method{}'s adaptive similarity/cardinality region. LOL-BO \citep{maus2022local} provides a complementary local-BO comparison in a learned latent space. In our experiments we use its pretrained SELFIES VAE without further representation updates while retaining the latent trust region, deep kernel surrogate, and Thompson sampling-based selection; we therefore denote this comparator as LOL-BO (frozen VAE).

All methods are evaluated under the same 1,000-call accounting and common initialization. Where a published method uses a different feedback or update frequency, we adapt its evaluation schedule to the common batched setting used here while retaining its underlying search strategy. This is important for interpreting the comparison, i.e., the experiment is designed to compare how these different search mechanisms use the same limited oracle budget, rather than to
reproduce each method under the larger or differently scheduled protocol used in its original publication. Additional implementation details for the comparators are given in Appendix~\ref{app:benchmark_summary}.

\begin{figure*}[t]
    \centering
    \includegraphics[width=\textwidth]{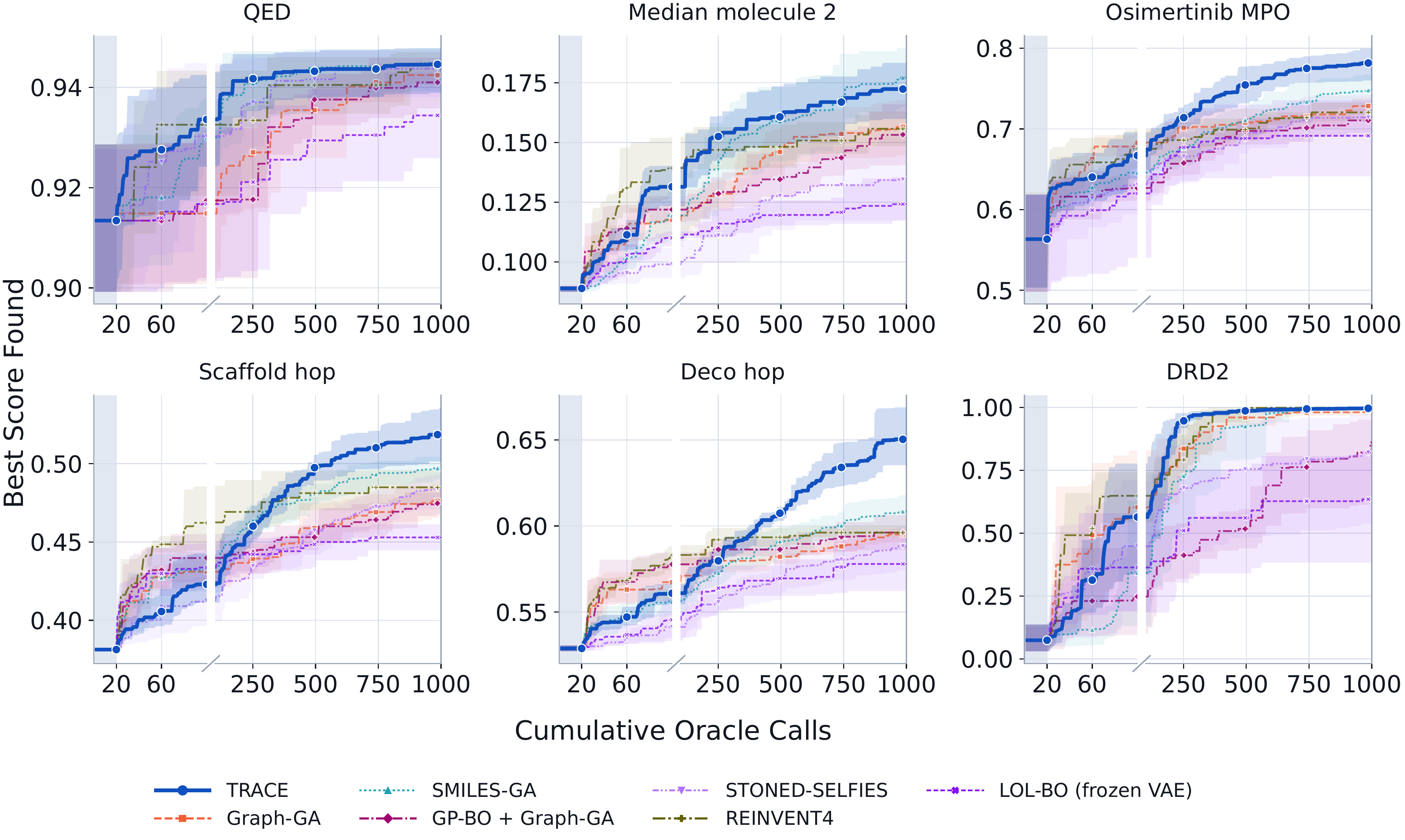}
    \caption{
    \textbf{Sample-efficient molecular optimization across six PMO objectives.}
    Mean best-so-far score over five seeds as a function of cumulative oracle evaluations, where the 1,000-call budget includes initialization. The gray
    region denotes the common 20-molecule initial set; active optimization then proceeds for 20 batches of $q=49$. The broken x-axis enlarges the early
    low-budget regime so that differences soon after initialization remain visible. Shaded bands are 95\% percentile-bootstrap intervals obtained by
    resampling whole runs. Appendix~\ref{app:benchmark_summary} summarizes the
    trajectories using top-1 and top-10 AUC.}
    \label{fig:benchmark_curves}
\end{figure*}

Figure~\ref{fig:benchmark_curves} shows a consistent advantage for \method{} once active optimization begins. Because every method starts from the same 20-molecule set, the subsequent separation of the trajectories reflects how the methods allocate the remaining oracle calls rather than differences in initialization. The clearest gains appear on median molecule 2, osimertinib MPO, scaffold hopping, and decoration hopping. On these objectives, \method{} continues to identify improvements later in the budget while several of the competing trajectories begin to flatten. This behavior is particularly
relevant to the motivation for adaptive search scale: progress is not confined to the first few batches, but continues as the region and candidate breadth are updated from the optimization history. QED is much less discriminating because the strongest methods rapidly approach the top of its score range, while DRD2 is the main case in which \method{} does not lead the best-so-far summary; REINVENT4 attains a slightly higher top-1 AUC.

Appendix~\ref{app:benchmark_summary} quantifies these trajectories using AUC summaries computed over the active portion of the budget. On top-1 AUC, which summarizes how quickly the best molecule improves, \method{} ranks first on four of the six objectives, matches SMILES-GA at the reported precision on QED, and ranks second on DRD2. Its mean top-1 AUC rank is $1.73$, the best among the compared methods. The sample-efficiency difference is also visible
directly from the trajectories: on osimertinib MPO, scaffold hopping, and decoration hopping, \method{} reaches approximately the full-budget best-so-far performance of the strongest comparator after only about half of the available oracle calls.

The conclusion is not specific to scoring each run by a single exceptional molecule. Appendix~\ref{app:benchmark_summary} also reports top-10 AUC, which tracks the mean score of the ten best molecules accumulated during a run and is closer to the aggregate measure emphasized in PMO. Under this metric, \method{} has the highest AUC on all six objectives and the best mean rank of $1.27$. Taken together, the best-so-far trajectories and the two AUC summaries
provide encouraging evidence that adapting the scale of candidate search can make more effective use of a small oracle budget (at least across these tests). 

\subsection{Adaptive region control depends on the proposal distribution}
\label{sec:region_ablation}

The benchmark trajectories in Figure~\ref{fig:benchmark_curves} establish the sample efficiency of TRACE under the low-budget protocol, but they do not separate the contribution of adaptive region control from that of the molecular proposal mechanism. We thus perform a crossed ablation over two proposal mechanisms and two region control settings:
\[
\{\text{TRACE proposer},\text{Graph-GA proposer}\}
\times
\{\text{adaptive region on},\text{adaptive region off}\}.
\]
The TRACE proposer is the objective-free graph-perturbation mechanism used throughout the method, while the Graph-GA proposer provides a population-based alternative. Within each paired comparison, the underlying proposal mechanism is held fixed and only the adaptive-region controller is enabled or disabled. When the region is disabled, $k$ is held at $k_0$ and neither the Tanimoto-similarity threshold nor the top-$k$ admission truncation is applied before acquisition optimization; Appendix~\ref{app:region_ablation} gives the complete protocol.

\begin{figure*}[t]
    \centering
    \includegraphics[width=0.95\textwidth]{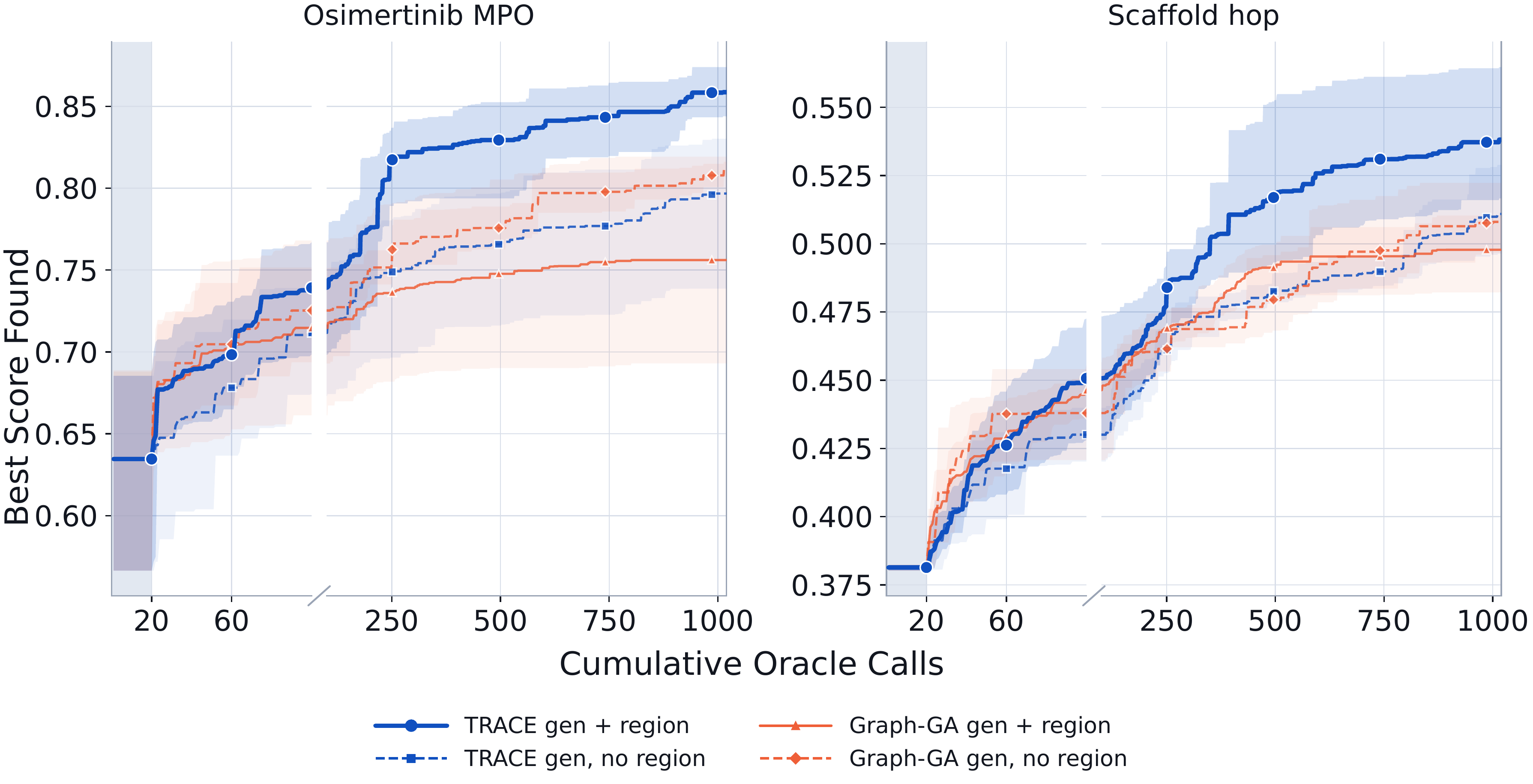}
    \caption{\textbf{Adaptive region control depends on the proposal distribution.}
    Best-so-far trajectories for osimertinib MPO and scaffold hopping under the TRACE and Graph-GA proposers, each with and without TRACE's adaptive region controller. All four configurations share the same $20$ evaluated initialization molecules for a given objective and seed. The gray region marks this common initialization, after which active optimization proceeds for $100$ iterations with $q=10$, ending at $1{,}020$ cumulative oracle evaluations. The broken x-axis magnifies the early trajectory only and does not alter oracle-call accounting. Curves show means over five seeds and shaded bands are $95\%$ percentile-bootstrap intervals obtained by resampling whole runs.}
    \label{fig:region_ablation}
\end{figure*}

For each objective and seed, all four configurations begin from the same $20$ evaluated initialization molecules and then perform $100$ active iterations with batch size $q=10$, for $1{,}020$ cumulative oracle evaluations. We evaluate the crossed design on osimertinib MPO and scaffold hopping using five seeds.
Figure~\ref{fig:region_ablation} shows a different effect of adaptive region control under the two proposal distributions. With the TRACE proposer, the region-enabled and region-disabled trajectories initially track one another but separate as optimization proceeds, with adaptive region control improving the mean best-so-far trajectory on both osimertinib MPO and scaffold hopping. Because the underlying proposal mechanism is unchanged within this pair, these comparisons provide evidence that the adaptive region contributes beyond the TRACE proposer itself on the two objectives studied.

The corresponding Graph-GA comparisons do not show the same benefit: enabling the adaptive region yields mildly lower mean trajectories on both objectives in these experiments. Thus, the controller should not be interpreted as a generic wrapper that necessarily improves an arbitrary proposal mechanism. Instead, the results are consistent with an interaction between adaptive locality and the candidate distribution supplied by the proposer. The TRACE proposer generates structured variation around the current search center $c_t$, after which the adaptive region controls how much of that variation is exposed to BO. The more dispersed Graph-GA proposal distribution may interact less favorably with the progressively localized admission rule. Controller success and restart diagnostics in Appendix~\ref{app:region_ablation} support this interpretation, while remaining descriptive rather than establishing a general mechanism for other proposal distributions.

\subsection{DFT-driven optimization of FePc dissolution resistance}
\label{sec:dft_case_study}

We next evaluate TRACE in a setting where the oracle is a first-principles electronic structure calculation rather than an inexpensive benchmark function. Iron phthalocyanine (FePc) has been widely studied as a non-precious molecular electrocatalyst for the oxygen reduction reaction (ORR) \citep{chen2020fepc,chen2019operando}. Its operation under acidic conditions, however, can be limited in part by degradation and demetalation of the Fe center \citep{baranton2005fepc,wan2022fepcdegradation}. Thus, we use a DFT-based dissolution energy objective to test whether TRACE can identify FePc-derived structures with greater computed thermodynamic resistance to Fe dissolution using a small number of high-fidelity evaluations. 

We optimize the Fe dissolution energy, $E_{\mathrm{diss}}$, using the thermodynamic construction of \citep{maraschin2026unraveling}:
\begin{equation}
E_{\mathrm{diss}}
=
E_{\mathrm{without~Fe}}
-
E_{\mathrm{with~Fe}}
+
\mu(\mathrm{Fe}^{2+}),
\label{eq:dft_dissolution}
\end{equation}
where $E_{\mathrm{with~Fe}}$ and $E_{\mathrm{without~Fe}}$ are the calculated energies of the corresponding Fe-containing and Fe-removed structures. Under this convention, $E_{\mathrm{diss}}$ is the energetic cost of removing the Fe center and transferring it to the Fe$^{2+}$ reservoir; a larger $E_{\mathrm{diss}}$ therefore corresponds to greater computed thermodynamic resistance to Fe dissolution. The Fe-containing and Fe-removed structures are evaluated using the same staged workflow: UFF relaxation \citep{rappe1992uff}, UMA pre-relaxation \citep{wood2025uma}, PBE geometry optimization \citep{perdew1996pbe}, and final HSE06 single-point energies \citep{heyd2003hse,krukau2006hse06}. Appendix~\ref{app:dft_details} provides the complete computational details.

The initial FePc reference ($E_{\mathrm{diss}}=0.42$~eV) is evaluated once before closed-loop optimization. Molecular generation is restricted to FePc-derived structures retaining the central Fe-N coordination core. At each round, TRACE generates proposals around the search center $c_t$, applies the adaptive $(\tau_t,k_t)$ region, and uses the fragprint Tanimoto-kernel GP with qLogEI \citep{ament2023unexpected} to select the DFT batch. Eight rounds with batch size $q=10$ yield 80 designed-candidate DFT evaluations in addition to the initial reference.

\begin{figure*}[t]
\centering
\begin{minipage}[t]{0.45\textwidth}
\centering
\includegraphics[width=\linewidth]{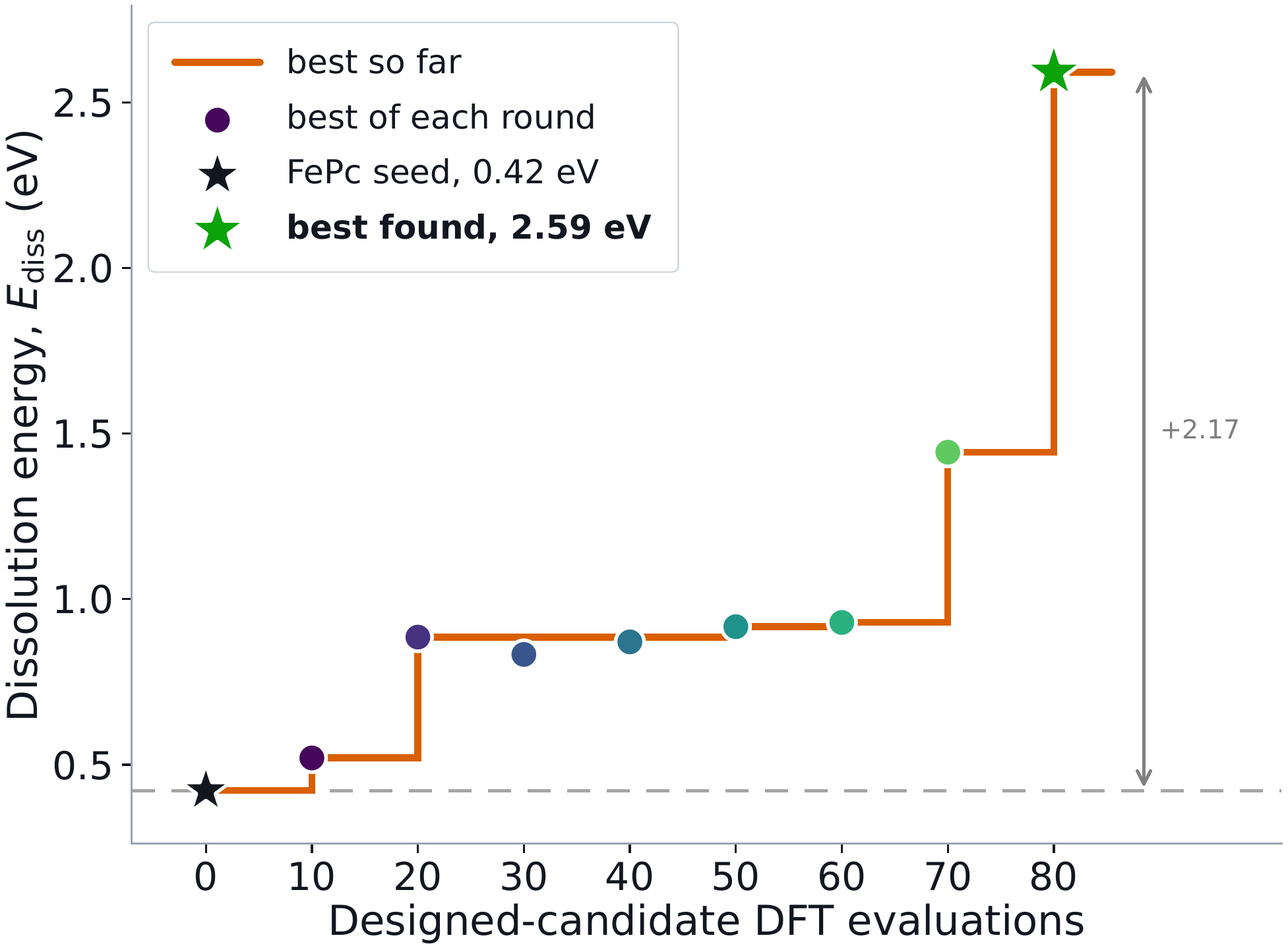}
\vspace{-2mm}

{\small (a)}
\end{minipage}
\hfill
\begin{minipage}[t]{0.45\textwidth}
\centering
\includegraphics[width=\linewidth]{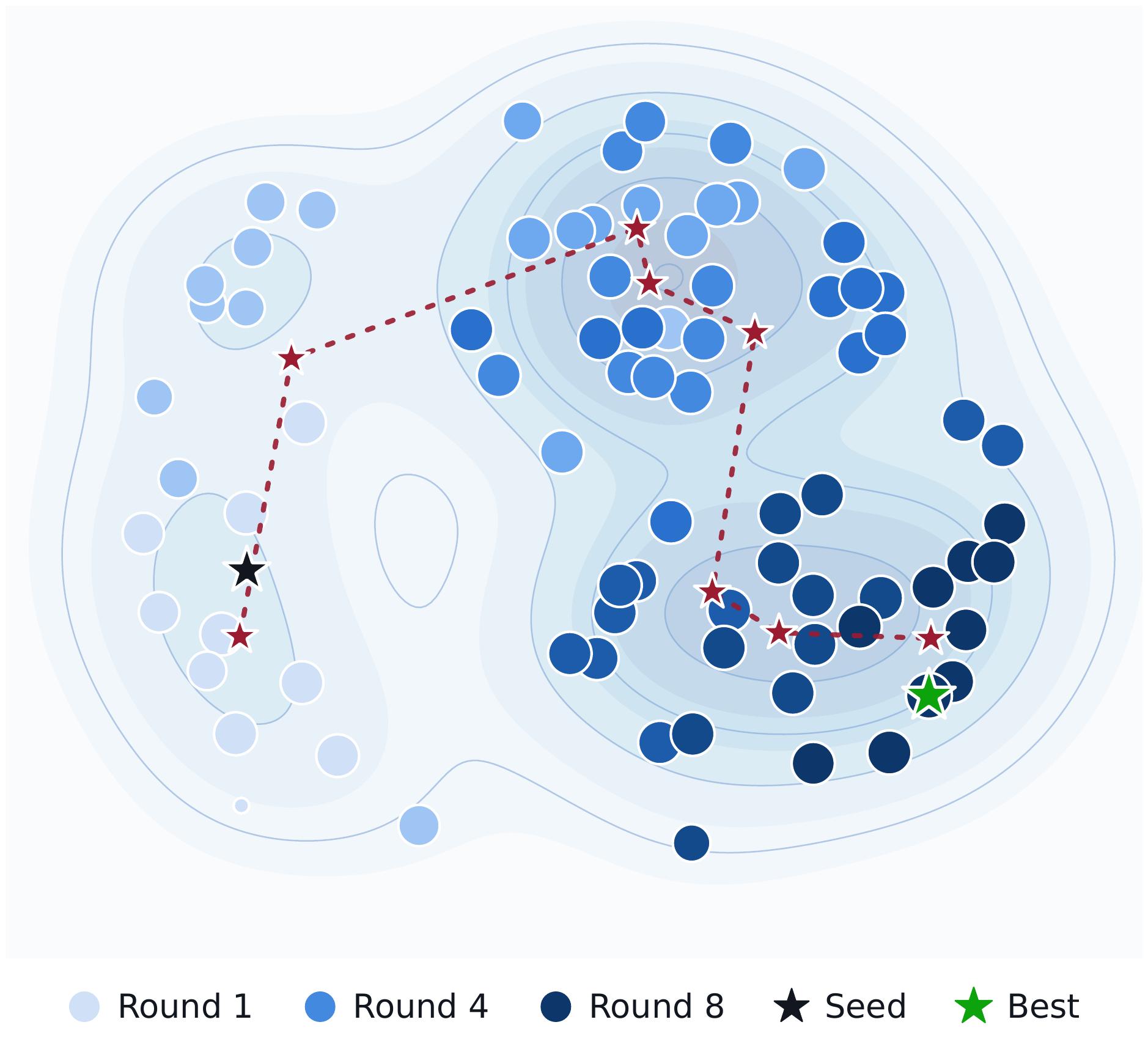}
\vspace{-2mm}

{\small (b)}
\end{minipage}

\caption{
\textbf{Closed-loop DFT optimization of FePc derivatives with TRACE.}
\textbf{(a)} Best-observed dissolution energy as a function of cumulative designed-candidate DFT evaluations. The initial FePc reference at $E_{\mathrm{diss}}=0.42$~eV is shown at zero designed evaluations and is not included in the subsequent $80$ designed-candidate evaluations. TRACE identifies a derivative with $E_{\mathrm{diss}}=2.59$~eV after those $80$ evaluations, an absolute increase of $2.17$~eV. \textbf{(b)} Two-dimensional multidimensional-scaling (MDS) visualization constructed from pairwise Morgan-fingerprint Tanimoto distances between the FePc seed and DFT-evaluated molecules. Color denotes optimization round, numbered markers indicate the centroid of each round in the two-dimensional embedding, and stars denote the initial FePc structure and the best molecule found. The projection is used only for qualitative visualization; quantitative similarity statistics in the original fingerprint space are reported in Appendix~\ref{app:dft_diagnostics}.}
\label{fig:dft_trace}
\end{figure*}

TRACE improves $E_{\mathrm{diss}}$ to approximately $0.52$~eV in round 1 and $0.88$~eV in round 2. The global incumbent remains near $0.9$~eV through the middle of the campaign, then reaches $1.44$~eV in round 7 and $2.59$~eV in round 8 (Figure~\ref{fig:dft_trace}a), a total increase of $2.17$~eV over the initial FePc reference.

Figure~\ref{fig:dft_trace}b provides a qualitative view of the movement through chemical space: early batches lie nearer FePc, while later rounds occupy different regions of the MDS embedding. The quantitative diagnostic in Appendix~\ref{app:dft_diagnostics} measures similarity in the original fingerprint space relative to the global incumbent $x_t^{\mathrm{best}}$, not the search center $c_t$. By round 8, the median similarity of evaluated molecules to the global incumbent is approximately $0.69$, compared with $0.19$ to the initial FePc seed. Thus, later candidates remain relatively close to the current best chemistry even as the campaign moves away from the starting molecule.

The final improvement demonstrates locality relative to TRACE's actual search center. The best round-8 structure has Tanimoto similarity $T=0.65$ to that round's center $c_t$, which has $E_{\mathrm{diss}}=1.44$~eV and is only approximately $T=0.19$ from the FePc seed. The resulting increase to $E_{\mathrm{diss}}=2.59$~eV is a further improvement of $1.15$~eV. This final transition illustrates the ``local steps, distant destination'' behavior that motivates TRACE: the improved molecule is local to the active search center, even though that center has moved far from the original FePc.

\section{Discussion and Limitations}
\label{sec:discussion}

The crossed ablation in Section \ref{sec:region_ablation} suggests that the adaptive region is most useful when considered together with the proposal mechanism on which it acts. With the TRACE proposer, turning ``on'' region adaptation improves performance on both objectives studied; applying the same controller to Graph-GA proposals does not. The diagnostics do not tell us exactly why, but they are consistent with the idea that changing the search region is only useful if the proposal mechanism can continue to supply appropriate candidates as that region changes. The pLogP experiment raises a related issue from another direction: what an optimizer can achieve also depends on the chemistry its generator makes accessible. These results make us hesitant to think of locality as a generic wrapper around an otherwise fixed optimizer. How candidates are generated and how the search region adapts appear to be closely connected design choices.

We therefore see the current TRACE implementation as one concrete instance of a broader adaptive-search idea. Morgan-fingerprint Tanimoto similarity provides a simple and interpretable definition of locality, but there is nothing fundamental about that particular choice, and fingerprint similarity is only a proxy for locality of the objective. The same is true of the linear coupling between $\tau_t$ and $k_t$. Other representations could define the neighborhood differently, while more flexible controllers could adapt its extent and candidate breadth separately or in a task-dependent way. There is a practical computational question here as well: broadening the candidate pool increases the work spent generating and ranking molecules. A more developed version of this idea could thus account for both oracle progress and the computational cost of searching a region when deciding how that region should change.

The FePc campaign is encouraging because this behavior carries over to a setting where the oracle is genuinely expensive. Over 80 designed-candidate DFT evaluations, TRACE moved well away from the initial FePc chemistry while improving computed resistance to Fe dissolution, and the final high-value transition remained local to the active search center. The present study is still one computational campaign on one objective, so practical catalyst design will require a broader treatment of competing objectives and constraints together with experimental validation. Because repeated DFT campaigns would be costly and a single run per method would be difficult to interpret, we treat this study as an expensive-oracle demonstration rather than a comparison of optimizers. Continuous-domain work has already shown that trust-region BO can be extended to constrained multi-objective problems \citep{daulton2022morbo}, providing one starting point for such an extension. Open-ended molecular design adds its own difficulty, however, because the region being adapted and the chemistry available inside it are both determined by the molecular representation and proposal process. Understanding that interaction more systematically is, in our view, an important next step.


\section{Conclusion}
\label{sec:conclusion}

We introduced TRACE, a Bayesian optimization framework that adapts the extent and breadth of molecular search as evaluations accumulate. Across six practical molecular optimization (PMO) benchmark objectives under matched 1,000-evaluation budgets, TRACE showed the strongest overall sample efficiency among the tested methods. Crossed ablations on two objectives showed that adaptive region control helped with TRACE proposals but not with Graph-GA proposals. In a closed-loop DFT campaign, TRACE identified an FePc derivative with $E_{\mathrm{diss}}=2.59$~eV, compared with $0.42$~eV for the starting molecule, after 80 designed-candidate evaluations. Together, these results support adapting chemical search scale under limited evaluation budgets.


\section*{Acknowledgments}

This work was supported by the National Science Foundation (NSF Award No. 2515046). We also thank the High Performance Computing Center at Texas Tech University for providing computational resources for the DFT calculations.

\bibliographystyle{unsrt}
\bibliography{TRACE_final_submission}

\newpage


\appendix

\newcommand{\reviewnote}[1]{%
  \par\smallskip
  \noindent
  \textcolor{red}{\textbf{[REVIEW NOTE:} #1\textbf{]}}
  \par\smallskip
}

\section{TRACE algorithmic details}
\label{app:algorithms}

\subsection{Outer-loop optimization}
\label{app:trace_outer}

Algorithm~\ref{alg:trace} expands the main-text description into the complete TRACE state transition using the notation of Section~\ref{sec:trace_design}. In addition, $s$ and $\ell$ are the current numbers of consecutive controller successes and failures. The fresh strict local pool is denoted $\mathcal{L}_t$, the persistent relaxed proposal reservoir by $\mathcal{G}_t$, and the conditional relaxed underfill pool by $\mathcal{F}_t$; their deduplicated union forms the proposal set $\mathcal{P}_t$ used by the admission rule.

To keep the pseudocode readable, we use four helper operations. $\textsc{RelaxedPool}(c,p)$ runs the objective-free generator in Algorithm~\ref{alg:gen} with the relaxed proposal mode around center $c$ and returns a finite proposal pool; $p\in\{\texttt{reservoir},\texttt{fallback}\}$ indicates whether the result is the persistent reservoir or a temporary underfill augmentation. $\textsc{Admit}(\mathcal{P},\mathcal{D},c,\tau,k)$ returns the admissible set defined by Eq.~\eqref{eq:budget}: it removes molecules already present in $\mathcal{D}$, applies the Tanimoto-similarity threshold, and retains at most the $k$ most similar survivors. $\textsc{SelectBatch}(\mathcal{A},\mathcal{M},q)$ returns $q$ molecules from $\mathcal{A}$ by optimizing the chosen batch acquisition function under GP posterior $\mathcal{M}$. Finally, $\textsc{RestartCenter}(\mathcal{D},c)$ returns an evaluated molecule to use as the new center, using the stored center-level optimization history and the restart heuristic described below. The notation $\operatorname{dedup}(\cdot)$ denotes removal of duplicate molecular structures.

\begin{algorithm}
\caption{\method{}: adaptive-region Bayesian optimization in chemical space}
\label{alg:trace}
\small
\begin{algorithmic}[1]
\Require evaluated initial data $\mathcal{D}_0$; total oracle budget $N$ including $\mathcal{D}_0$; nominal batch size $q$
\Require controller parameters $k_{\min},k_{\max},k_0$; coupling $\tau(k)$ from Eq.~\eqref{eq:coupling}; thresholds $n_{\mathrm{succ}},n_{\mathrm{fail}}$; tolerance $\varepsilon$
\Ensure global incumbent after exhausting the oracle budget
\State $t\gets0$; $k_t\gets k_0$; $s,\ell\gets0,0$ \Comment{success/failure counters}
\State $x_t^{\mathrm{best}}\gets\operatorname*{arg\,max}_{x\in\mathcal{X}(\mathcal{D}_t)}f(x)$; $c_t\gets x_t^{\mathrm{best}}$
\State $\mathcal{G}_t\gets\textsc{RelaxedPool}(c_t,\texttt{reservoir})$
\While{$|\mathcal{D}_t|<N$}
\Statex \textit{Generate and admit candidates}
\State $q_t\gets\min(q,N-|\mathcal{D}_t|)$; $\tau_t\gets\tau(k_t)$
\State $\mathcal{L}_t\gets\textsc{Propose}(c_t,\max(50,k_t),\texttt{strict})$ \Comment{Algorithm~\ref{alg:gen}}
\State $\mathcal{P}_t\gets\operatorname{dedup}(\mathcal{G}_t\cup\mathcal{L}_t)$
\State $\mathcal{A}_t\gets\textsc{Admit}(\mathcal{P}_t,\mathcal{D}_t,c_t,\tau_t,k_t)$
\If{$|\mathcal{A}_t|<q_t$}
\State $\mathcal{F}_t\gets\textsc{RelaxedPool}(c_t,\texttt{fallback})$
\State $\mathcal{P}_t\gets\operatorname{dedup}(\mathcal{P}_t\cup\mathcal{F}_t)$; $\mathcal{A}_t\gets\textsc{Admit}(\mathcal{P}_t,\mathcal{D}_t,c_t,\tau_t,k_t)$
\EndIf
\If{$|\mathcal{A}_t|<q_t$}
\State $c_t\gets\textsc{RestartCenter}(\mathcal{D}_t,c_t)$; $k_t\gets k_0$; $s,\ell\gets0,0$
\State $\mathcal{G}_t\gets\textsc{RelaxedPool}(c_t,\texttt{reservoir})$; \textbf{continue}
\EndIf
\Statex \textit{Select and evaluate a batch}
\State fit the fragprint Tanimoto-kernel GP to $\mathcal{D}_t$ and denote its posterior by $\mathcal{M}_t$
\State $B_t\gets\textsc{SelectBatch}(\mathcal{A}_t,\mathcal{M}_t,q_t)$
\State evaluate $f$ on $B_t$ and set $\mathcal{D}_{t+1}\gets\mathcal{D}_t\cup\{(x,f(x)):x\in B_t\}$
\State $x_{t+1}^{\mathrm{best}}\gets\operatorname*{arg\,max}_{x\in\mathcal{X}(\mathcal{D}_{t+1})}f(x)$
\State $c_{t+1}\gets x_{t+1}^{\mathrm{best}}$; $k_{t+1}\gets k_t$; $\mathcal{G}_{t+1}\gets\mathcal{G}_t$
\Statex \textit{Update the adaptive region}
\If{$f(x_{t+1}^{\mathrm{best}})>f(x_t^{\mathrm{best}})+\varepsilon\max(1,|f(x_t^{\mathrm{best}})|)$}
\State $s\gets s+1$; $\ell\gets0$
\Else
\State $\ell\gets\ell+1$; $s\gets0$
\EndIf
\If{$s\ge n_{\mathrm{succ}}$}
\State $k_{t+1}\gets\min(k_{\max},2k_t)$; $s\gets0$
\ElsIf{$\ell\ge n_{\mathrm{fail}}$}
\If{$k_t/2<\max(k_{\min},q_t)$}
\State $c_{t+1}\gets\textsc{RestartCenter}(\mathcal{D}_{t+1},c_{t+1})$; $k_{t+1}\gets k_0$; $s,\ell\gets0,0$
\State $\mathcal{G}_{t+1}\gets\textsc{RelaxedPool}(c_{t+1},\texttt{reservoir})$
\Else
\State $k_{t+1}\gets k_t/2$; $\ell\gets0$
\EndIf
\EndIf
\State $t\gets t+1$
\EndWhile
\State \Return $x_t^{\mathrm{best}}$
\end{algorithmic}
\end{algorithm}

\paragraph{Controller and restart semantics.} The incumbent and the ordinary next search center are updated after every batch, even when the improvement is smaller than $\varepsilon\max(1,|f(x_t^{\mathrm{best}})|)$ and therefore counts as a controller failure. Expansion and contraction change only the adaptive state $k$; the corresponding similarity threshold is obtained from Eq.~\eqref{eq:coupling} at the next iteration. Proposal generation, admission, fallback generation, and a restart occurring before batch selection do not consume oracle evaluations.

$\textsc{RestartCenter}$ does not modify the global incumbent. It first deduplicates previously used iteration centers and considers them in descending order of center objective value, best improvement obtained from that center, and maximum admissible-pool size observed from that center, subject to a Morgan-Tanimoto screening rule against earlier restart centers. If no prior center qualifies, it considers all evaluated molecules in descending objective-value order under the same screening rule; the global incumbent is the final fallback. The current center is excluded from the first two search passes.


\subsection{Objective-free proposal generation}
\label{app:proposal_algorithm}

Algorithm~\ref{alg:gen} defines the generator used for the strict local pool and both relaxed pools in Algorithm~\ref{alg:trace}. Its inputs are the search center $c$, requested target size $M$, proposal mode $m\in\{\texttt{strict},\texttt{relaxed}\}$, graph-edit operator set $\mathcal{O}$, and center-derived constraint set $\mathcal{K}_m(c)$. The accepted proposal library is $\mathcal{L}$; $x$ denotes the current molecule from which edits are proposed and $y$ a newly generated candidate. All internal adaptation uses structural or proposal-history information only.

\begin{algorithm}[H]
\caption{\textsc{Propose}: objective-free molecular generation}
\label{alg:gen}
\small
\begin{algorithmic}[1]
\Require search center $c$; target size $M$; mode $m\in\{\texttt{strict},\texttt{relaxed}\}$; graph-edit operators $\mathcal{O}$; center-derived constraints $\mathcal{K}_m(c)$
\Ensure proposal library $\mathcal{L}$ with $|\mathcal{L}|\le M$
\State $\mathcal{L}\gets\emptyset$; $x\gets c$; initialize objective-free proposal statistics for mode $m$
\While{$|\mathcal{L}|<M$ and the objective-free stopping criterion is not met}
\State $y\gets\textsc{EditChain}(x,\mathcal{O},m)$ using the current proposal statistics
\If{$y$ is invalid, duplicated, or violates $\mathcal{K}_m(c)$}
\State record the rejected outcome and \textbf{continue}
\EndIf
\State compute structural statistics for $y$; add $y$ to $\mathcal{L}$; update proposal/operator statistics
\State choose the next edit seed $x$ from $c$ and accepted proposals using the updated objective-free proposal state
\EndWhile
\State \Return $\mathcal{L}$
\end{algorithmic}
\end{algorithm}

\paragraph{Edits, constraints, and stopping.} $\textsc{EditChain}(x,\mathcal{O},m)$ applies one or more chemistry-aware graph edits to the current edit seed $x$ and returns a candidate molecule $y$. The operator set includes atom and bond mutations, fragment addition and deletion, and ring fuse, bridge, spiro, open, shrink, and peel operations. Invalid and duplicate molecules are rejected. The internal locality, shell-coverage, novelty, and complexity statistics use the Morgan-fingerprint similarity $T$ defined in Section~\ref{sec:trace_design}; they may alter operator use and edit-seed selection but never use oracle values. The objective-free stopping criterion permits the generator to return fewer than $M$ molecules when valid or novel proposal generation saturates; its numerical thresholds and rejection limits are left to the implementation.

The mode $m$ selects both the proposal behavior and one of two center-derived constraint profiles. The \texttt{strict} mode is used for the fresh local pool $\mathcal{L}_t$ generated every TRACE iteration. The \texttt{relaxed} mode is used for the persistent reservoir $\mathcal{G}_t$ generated at initialization and after restarts and for the conditional fallback pool $\mathcal{F}_t$. Each constraint set $\mathcal{K}_m(c)$ is re-derived from the center used to generate that pool. For the strict local pool, TRACE requests $M=\max(50,k_t)$ proposals. Exact constraint bounds, relaxed-pool target sizes, operator probabilities, proposal-controller constants, and other low-level generation settings are left to the implementation.

No oracle value enters Algorithm~\ref{alg:gen}. Oracle information first enters the proposal-to-evaluation pipeline through the GP/acquisition stage in Algorithm~\ref{alg:trace}; the outer TRACE loop also uses oracle values to update the incumbent, choose ordinary search centers, drive the success/failure controller, and inform the restart history.

\section{Low-budget benchmark details and summary statistics}
\label{app:benchmark_summary}

\subsection{Benchmark implementation}
\label{app:benchmark_implementation}

The benchmark in Section~\ref{sec:low_budget_benchmarks} uses a common
evaluation contract across all methods. For each objective and seed, the methods begin from the same $20$ evaluated molecules and then perform $20$ active rounds of $49$ oracle evaluations. The same PMO oracle and molecular filter (SA $\leq 7$; SMILES length $\leq 100$) are used throughout. This common initialization is also the starting point of the trajectories in Figure~\ref{fig:benchmark_curves}.

For \method{}, we use the same controller settings for all six objectives: $\tau_{\mathrm{lo}}=0.10$, $\tau_{\mathrm{hi}}=0.50$,
$k_0=200$, $k_{\min}=20$, $k_{\max}=1000$,
$n_{\mathrm{succ}}=5$, $n_{\mathrm{fail}}=3$, and
$\varepsilon=10^{-3}$. The GP and adaptive region are updated after each evaluated batch, and qLogEI is optimized over the resulting admissible set. No objective-specific tuning of these controller settings is performed.

The comparators retain the search mechanisms described in their cited references, but their evaluation schedules are placed under the same $20$-round, $1{,}000$-call contract so that differences in the trajectories reflect how the methods use a common oracle budget. This requires a benchmark-specific change for methods whose original implementations use a different feedback frequency. For example, GP-BO + Graph-GA \citep{tripp2024diagnosing} retains the GP-based BO and Graph-GA acquisition search, but is updated according to the matched batched schedule used here.

The latent-space comparator also differs in one material respect from the full LOL-BO procedure \citep{maus2022local}. We keep the pretrained SELFIES VAE fixed throughout optimization while retaining the latent trust region, deep kernel sparse variational GP surrogate, and Thompson sampling-based selection. We therefore refer to this configuration throughout as LOL-BO (frozen VAE). For Graph-GA, SMILES-GA, STONED-SELFIES, and REINVENT4, the main text summarizes the distinct search mechanisms that motivate their inclusion. Exact software settings and implementation-level details for all comparator runs will be provided with the released code at \url{https://github.com/PaulsonLab/TRACE-AI4MAT}.

\subsection{Top-1 and top-10 AUC}
\label{app:benchmark_auc}

Figure~\ref{fig:benchmark_curves} shows the best molecule found as the oracle budget accumulates. We summarize these trajectories with AUC statistics while respecting the batched nature of the experiment. After the common initialization, observations arrive in batches of $49$, giving cumulative batch endpoints
\[
n_j = 20 + 49j, \qquad j=0,\ldots,20,
\]
with $n_0=20$ and $n_{20}=1000$. We do not impose an artificial ordering on the molecules within a batch. Instead, the relevant run-level statistic is updated when the complete batch is observed and held fixed until the next batch endpoint.

More generally, let $s_K(n)$ denote the mean score of the $K$ best molecules observed by cumulative oracle call $n$, with $K=1$ corresponding to the best-so-far trajectory and $K=10$ to the mean of the ten best molecules. We compute
\[
\mathrm{AUC}_{K}
=
\frac{1}{980}
\int_{20}^{1000} s_K(n)\,dn,
\qquad K\in\{1,10\}.
\]
The common $20$-molecule initialization therefore sets the starting value of each curve but is not itself integrated; the normalization is the $980$-call active interval. Because we normalize only by the number of oracle calls, the reported AUC values remain in the units of the underlying objective score.

Top-1 AUC directly summarizes the best-so-far behavior shown in Figure~\ref{fig:benchmark_curves}. Table~\ref{tab:auc} additionally reports the rank of each method by its final best molecule. The rank columns are computed by ranking the methods within each of the $30$ objective-seed combinations ($6$ objectives $\times$ $5$ seeds) and then averaging those ranks.

\begin{table}[t]
  \centering
  \caption{\textbf{Top-1 AUC and final-best performance.}
  Entries report mean top-1 AUC over five seeds under the low-budget protocol   described above. Bold marks the highest mean at the reported precision.   The two rank columns average each method's rank over the $30$ objective-seed combinations using top-1 AUC and the final best-so-far score, respectively; rows are ordered by mean AUC rank.}
  \label{tab:auc}
  \footnotesize
  \setlength{\tabcolsep}{4.0pt}
  \begin{tabular}{lcccccc@{\hskip 7pt}c@{\hskip 5pt}c}
    \toprule
    Method & QED & Median 2 & Osim.\ MPO & Scaf.\ hop & Deco hop & DRD2 & Rank (AUC) & Rank (final) \\
    \midrule
    \method{}            & \textbf{0.940} & \textbf{0.154} & \textbf{0.732} & \textbf{0.479} & \textbf{0.603} & 0.870 & \textbf{1.73} & \textbf{1.77} \\
    SMILES-GA            & \textbf{0.940} & 0.151 & 0.699 & 0.471 & 0.584 & 0.796 & 2.93 & 2.33 \\
    REINVENT4            & 0.938 & 0.146 & 0.694 & 0.472 & 0.589 & \textbf{0.874} & 3.07 & 3.97 \\
    Graph-GA             & 0.933 & 0.139 & 0.699 & 0.451 & 0.580 & 0.848 & 4.33 & 4.20 \\
    STONED-SELFIES       & 0.939 & 0.120 & 0.679 & 0.450 & 0.566 & 0.676 & 4.97 & 4.53 \\
    GP-BO + Graph-GA     & 0.931 & 0.133 & 0.674 & 0.452 & 0.584 & 0.548 & 5.03 & 5.00 \\
    LOL-BO (frozen VAE)  & 0.926 & 0.117 & 0.671 & 0.443 & 0.565 & 0.526 & 5.93 & 6.20 \\
    \bottomrule
  \end{tabular}
\end{table}

The top-1 summary mirrors the qualitative behavior in
Figure~\ref{fig:benchmark_curves}. \method{} has the best mean AUC rank ($1.73$) and final-best rank ($1.77$), leading outright on median molecule 2, osimertinib MPO, scaffold hopping, and decoration hopping. QED is close to saturation, with \method{} and SMILES-GA equal at the reported three-decimal precision. DRD2 is the exception to the otherwise leading top-1 pattern: REINVENT4 obtains an AUC of $0.874$, compared with $0.870$ for \method{}.

Top-1 AUC emphasizes discovery of the single-best molecule. To test whether the same conclusion holds when performance depends on accumulating several strong candidates, we also compute top-10 AUC using the same runs and the same batch-endpoint convention. This mirrors the top-10 within-run summary emphasized by PMO \citep{gao2022sample}, but here it is evaluated over our
$20$-$1{,}000$ low-budget interval and reported directly in the original objective-score units.

\begin{table}[t]
  \centering
  \caption{\textbf{Top-10 AUC.}
  Area under the mean-of-ten-best trajectory over the active $980$-call interval. Entries are means over five seeds, and bold marks the highest mean at the reported precision. The final column averages the top-10 AUC rank over the $30$ objective-seed combinations.}
  \label{tab:auc10}
  \footnotesize
  \setlength{\tabcolsep}{4.0pt}
  \begin{tabular}{lcccccc@{\hskip 7pt}c}
    \toprule
    Method & QED & Median 2 & Osim.\ MPO & Scaf.\ hop & Deco hop & DRD2 & Rank (top-10) \\
    \midrule
    \method{}            & \textbf{0.932} & \textbf{0.148} & \textbf{0.707} & \textbf{0.474} & \textbf{0.598} & \textbf{0.834} & \textbf{1.27} \\
    SMILES-GA            & 0.928 & 0.142 & 0.671 & 0.464 & 0.579 & 0.727 & 2.43 \\
    REINVENT4            & 0.906 & 0.129 & 0.638 & 0.449 & 0.576 & 0.637 & 4.00 \\
    STONED-SELFIES       & 0.926 & 0.111 & 0.651 & 0.442 & 0.562 & 0.635 & 4.20 \\
    Graph-GA             & 0.902 & 0.128 & 0.661 & 0.433 & 0.566 & 0.719 & 4.57 \\
    GP-BO + Graph-GA     & 0.896 & 0.122 & 0.628 & 0.432 & 0.571 & 0.344 & 5.33 \\
    LOL-BO (frozen VAE)  & 0.898 & 0.107 & 0.596 & 0.430 & 0.552 & 0.294 & 6.20 \\
    \bottomrule
  \end{tabular}
\end{table}

The top-10 results strengthen the aggregate picture without changing its interpretation. \method{} has the highest mean top-10 AUC on all six objectives and the best mean rank, $1.27$. The distinction is particularly clear on DRD2: although REINVENT4 has the slightly higher top-1 AUC, \method{} has the higher top-10 AUC ($0.834$ versus $0.637$). Thus, the single-best trajectory on DRD2 does not translate into the strongest ten-molecule summary. More broadly, the agreement between the top-1 and top-10 analyses indicates that the sample-efficiency gains in this six-objective study are not driven only by isolated high-scoring molecules.

\section{FePc DFT case study: computational details and diagnostics}
\label{app:dft_details}

\subsection{Electronic structure workflow}
\label{app:dft_workflow}

Each molecule selected by TRACE is represented initially as a SMILES string and converted into a three-dimensional molecular structure using RDKit \citep{rdkit}. The generated geometry is first optimized using the Universal Force Field (UFF) \citep{rappe1992uff}. We subsequently apply a machine-learning pre-relaxation using the Universal Model for Atoms (UMA), specifically the UMA-medium model with the \texttt{omol} task \citep{wood2025uma}. The \texttt{omol} task targets molecular electronic structure data represented in the Open Molecules 2025 (OMol25) dataset, which contains more than $10^8$ density functional theory (DFT) calculations at the $\omega$B97M-V/def2-TZVPD level of theory \citep{levine2025omol25}. UMA is used only to improve the geometry supplied to the subsequent first-principles calculations and does not provide the objective value used by TRACE.

Following UMA pre-relaxation, the structures are optimized using the Perdew-Burke-Ernzerhof (PBE) generalized-gradient approximation \citep{perdew1996pbe} as implemented in the Vienna \textit{Ab initio} Simulation Package (VASP) \citep{kresse1996vasp}. Final energies used in the dissolution calculation are obtained from single-point HSE06 hybrid-functional calculations \citep{heyd2003hse,krukau2006hse06} performed on the PBE-relaxed structures. The same staged procedure is applied to the Fe-containing and corresponding Fe-removed structures entering the dissolution-energy calculation. This workflow avoids the substantially greater cost of full hybrid-functional geometry optimization for every state; the use of hybrid-functional single-point energies on semilocal-DFT-relaxed geometries is a computational approximation whose accuracy can remain system dependent \citep{gouveia2019hybrid}.

\subsection{Dissolution-energy construction}
\label{app:dft_thermodynamics}

We use the metal-dissolution thermodynamic construction of \citep{maraschin2026unraveling}. For each FePc derivative,
\begin{equation}
E_{\mathrm{diss}}
=
E_{\mathrm{without~Fe}}
-
E_{\mathrm{with~Fe}}
+
\mu(\mathrm{Fe}^{2+}),
\label{eq:dissolution_appendix}
\end{equation}
where $E_{\mathrm{with~Fe}}$ and $E_{\mathrm{without~Fe}}$ are the calculated energies of the corresponding Fe-containing and Fe-removed structures. Under this convention, $E_{\mathrm{diss}}$ is the energetic cost of removing the Fe center and transferring it to the Fe$^{2+}$ reservoir; a larger $E_{\mathrm{diss}}$ therefore corresponds to greater computed thermodynamic resistance to Fe dissolution.

Rather than directly representing a solvated Fe$^{2+}$ ion with DFT, its electrochemical potential is referenced to bulk metallic Fe using the Fe$^{2+}$/Fe$^0$ redox equilibrium,
\begin{equation}
\mu(\mathrm{Fe}^{2+})
=
\mu(\mathrm{Fe(s)})
-
2e\left(U_{\mathrm{SHE}}-U_0\right),
\label{eq:fe_chemical_potential_appendix}
\end{equation}
where $\mu(\mathrm{Fe(s)})$ is the chemical potential of bulk metallic Fe, $U_{\mathrm{SHE}}$ is the applied potential relative to the standard hydrogen electrode, and $U_0$ is the equilibrium potential of the Fe$^{2+}$/Fe$^0$ redox couple. This construction avoids direct DFT treatment of the solvated Fe$^{2+}$ ion by referencing its electrochemical potential to bulk Fe and is analogous in spirit to the computational hydrogen electrode treatment of electrochemical thermodynamics \citep{norskov2004orr}.

\subsection{TRACE settings for the DFT campaign}
\label{app:dft_trace_settings}

The FePc reference is evaluated once before optimization starts, giving the initial value $E_{\mathrm{diss}}=0.42$~eV. TRACE then performs eight optimization rounds with batch size $q=10$, giving $80$ designed-candidate DFT evaluations after this initial reference evaluation. For this campaign, molecular generation is restricted to FePc-derived structures retaining the central Fe-N coordination core.

At round $t$, proposals are generated around the current search center $c_t$ using the objective-free graph-perturbation machinery described in Section~\ref{subsec:proposal}. The adaptive region is controlled by the Tanimoto-similarity threshold $\tau_t$ and cardinality budget $k_t$. The fragprint Tanimoto-kernel GP is fitted to the accumulated evaluated data, and qLogEI \citep{ament2023unexpected} is optimized over the admissible set to select the molecules submitted to the DFT oracle. The resulting dissolution energies are added to the evaluated data before the next round. The DFT campaign uses the controller settings in Table~\ref{tab:dft_trace_settings}.

\begin{table}[t]
\centering
\caption{TRACE settings used in the FePc DFT campaign.}
\label{tab:dft_trace_settings}
\begin{tabular}{lc}
\toprule
Parameter & Value \\
\midrule
Batch size $q$ & 10 \\
Number of rounds & 8 \\
$k_0$ & $2{,}000$ \\
$k_{\min}$ & 20 \\
$k_{\max}$ & $10{,}000$ \\
$\tau_{\mathrm{lo}}$ & 0.35 \\
$\tau_{\mathrm{hi}}$ & 0.70 \\
$n_{\mathrm{succ}}$ & 3 \\
$n_{\mathrm{fail}}$ & 2 \\
$\varepsilon$ & $1\times10^{-3}$ \\
Acquisition function & qLogEI \\
\bottomrule
\end{tabular}
\end{table}

\subsection{Chemical space dynamics}
\label{app:dft_diagnostics}

\begin{figure}[t]
\centering
\includegraphics[width=0.88\linewidth]{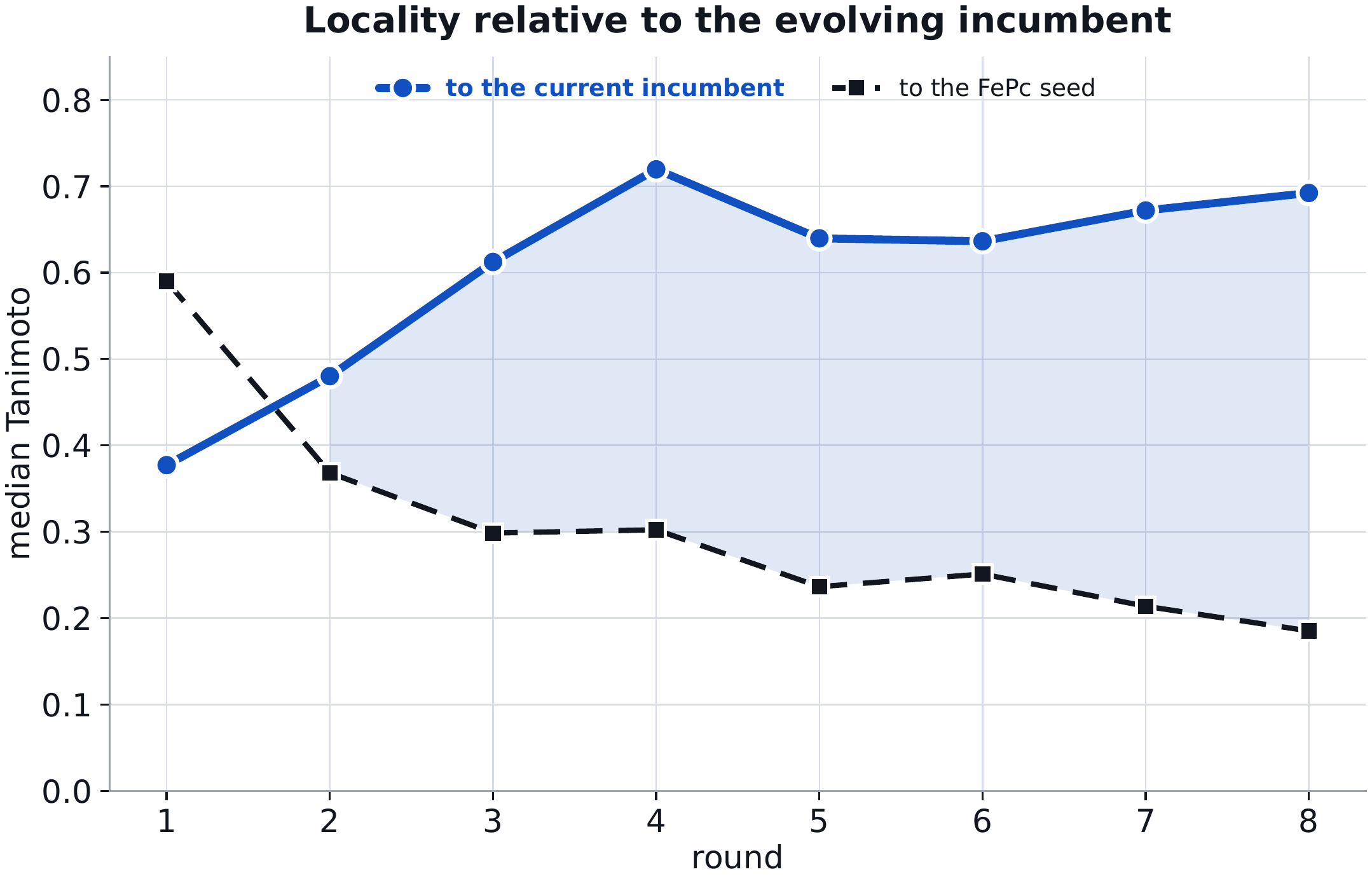}
\caption{
\textbf{Locality relative to the evolving incumbent and displacement from the FePc seed.}
Median Tanimoto similarity of the DFT-evaluated molecules in each round to the corresponding global incumbent $x_t^{\mathrm{best}}$ and to the original FePc seed. Similarity to the incumbent increases over the campaign, reaching approximately $0.69$ in round 8, while similarity to the original FePc molecule decreases to approximately $0.19$. The diagnostic therefore shows that later evaluated molecules remain relatively close to the current best chemistry while the campaign as a whole becomes substantially displaced from its starting point. Because the plotted comparison is incumbent-based, it is not interpreted as a direct measurement of similarity to the search center $c_t$ when the two may differ.}
\label{fig:dft_locality}
\end{figure}

To quantify cumulative structural displacement during the DFT campaign, we compute Tanimoto similarity using the radius-2, 2048-bit binary Morgan fingerprints defined in Section~\ref{sec:trace_design}. For each optimization round, Figure~\ref{fig:dft_locality} reports the median similarity of the DFT-evaluated molecules to both (i) the global incumbent $x_t^{\mathrm{best}}$ used in the round-wise diagnostic and (ii) the original FePc seed.

The two trends separate clearly as optimization proceeds. The median similarity to the initial FePc seed decreases from approximately $0.59$ in round 1 to approximately $0.19$ in round 8, whereas the median similarity to the round-specific global incumbent rises to approximately $0.69$. This incumbent-based diagnostic supports the cumulative-displacement aspect of the TRACE trajectory without requiring the two-dimensional MDS projection in Figure~\ref{fig:dft_trace}b to be interpreted quantitatively.

\subsection{Final transition around the evolved search center}
\label{app:dft_final_improvement}

\begin{figure}[t]
\centering
\includegraphics[width=\linewidth]{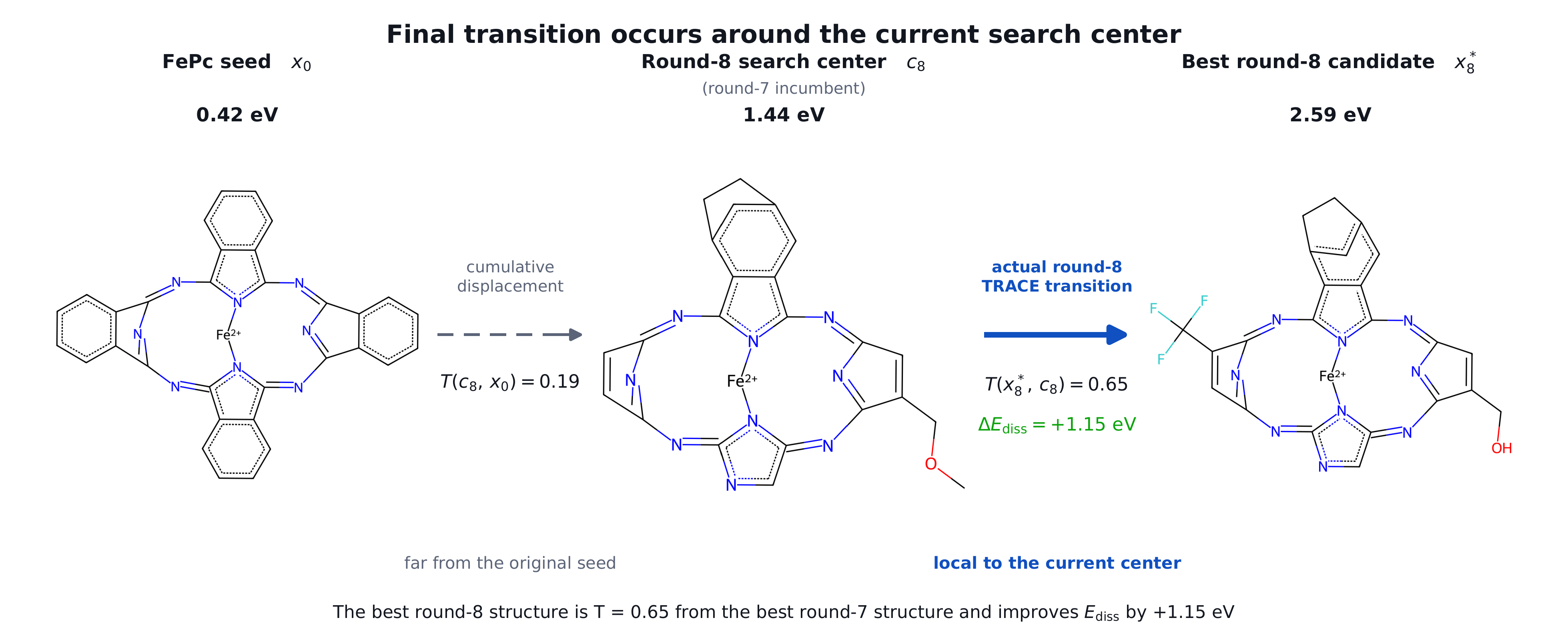}
\caption{
\textbf{Final improvement around the evolved TRACE search center.}
The FePc seed $x_0$, the round-8 search center $c_8$ (the round-7 incumbent), and the best molecule evaluated in round 8 are shown with their computed dissolution energies. By the start of round 8, the search center has moved substantially from the initial FePc structure, with $T(c_8,x_0)\approx0.19$. Round 8 is nevertheless conducted relative to this evolved center, and the best evaluated candidate has $T(x_8^\star,c_8)=0.65$, increasing $E_{\mathrm{diss}}$ from $1.44$ to $2.59$~eV. The two similarities therefore describe distinct quantities: cumulative displacement from the initial chemistry and
locality relative to the search center used for the final round.
}
\label{fig:dft_lineage}
\end{figure}

Figure~\ref{fig:dft_lineage} isolates the final transition in the closed-loop campaign and distinguishes locality with respect to the current TRACE search center from similarity to the initial molecule. At the start of round 8, the search center $c_8$ is the round-7 incumbent, with $E_{\mathrm{diss}}=1.44$~eV. Although this center has
moved substantially from the initial FePc seed
($T(c_8,x_0)\approx0.19$), round-8 proposals are generated and
filtered relative to $c_8$, rather than relative to $x_0$.

The best molecule evaluated in round 8 has $T(x_8^\star,c_8)=0.65$ and $E_{\mathrm{diss}}=2.59$~eV, corresponding to a $1.15$~eV increase over the preceding incumbent. Thus, the final improvement does not require a direct jump from the original FePc structure: it is obtained
from the neighborhood of an evolved search center that has itself become distant from the starting chemistry. This example illustrates how repeated re-centering can accumulate substantial global displacement while exploration remains referenced to the current center. 

\subsection{Round-wise molecular evolution}
\label{app:dft_roundwise}

\begin{figure}[t]
\centering
\includegraphics[width=0.95\linewidth]{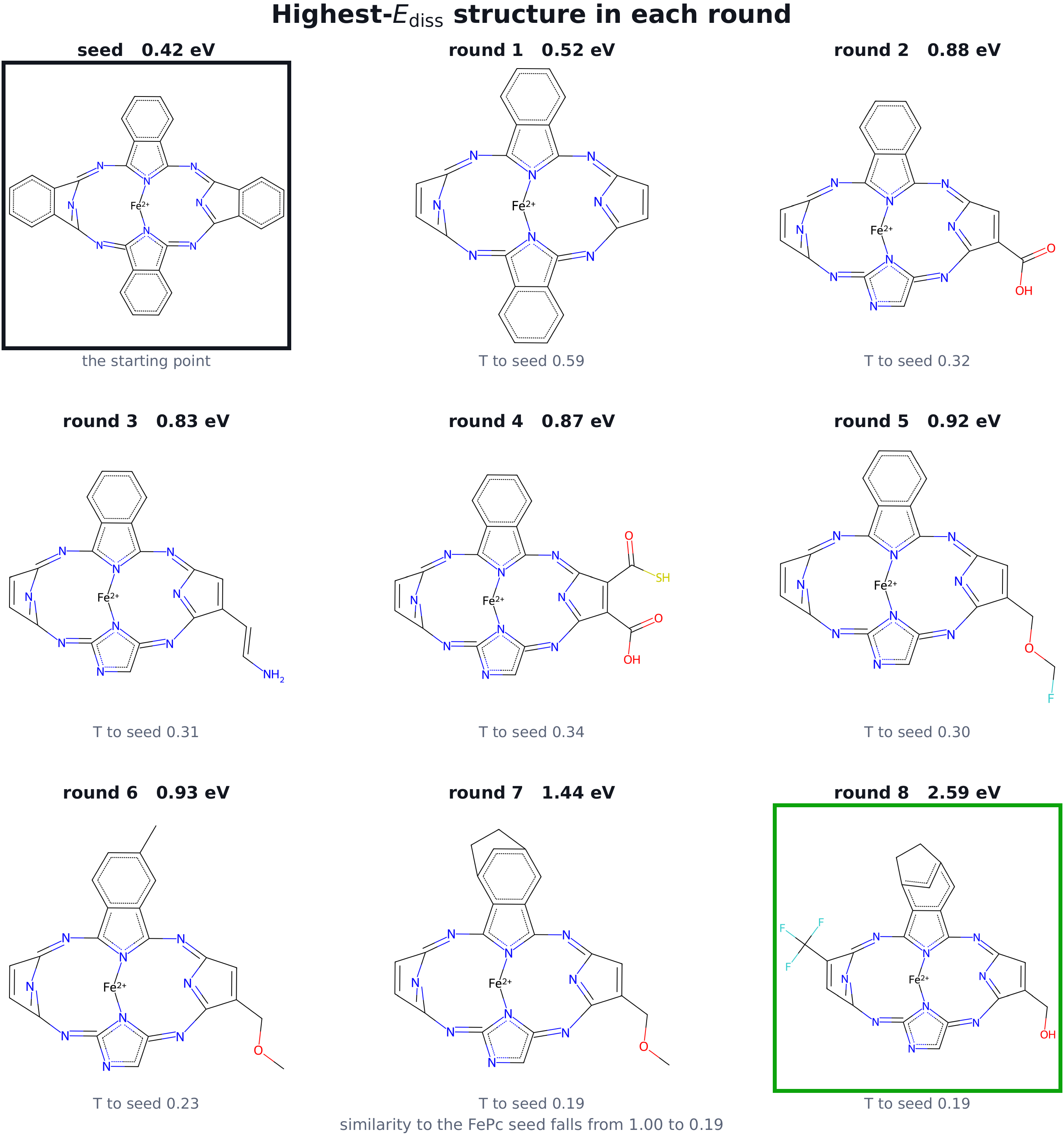}
\caption{
\textbf{Highest-$E_{\mathrm{diss}}$ DFT-evaluated FePc-derived structure in each TRACE round.}
Each panel shows the highest-$E_{\mathrm{diss}}$ molecule observed during that round together with its dissolution energy and Tanimoto similarity to the original FePc structure. The sequence illustrates the cumulative structural evolution of the search but is not used to infer a causal structure-property relationship or catalytic-performance improvement.}
\label{fig:dft_round_gallery}
\end{figure}

For completeness, Figure~\ref{fig:dft_round_gallery} shows the highest-$E_{\mathrm{diss}}$ molecule identified in each optimization round. The gallery provides a qualitative chemical audit of the trajectory and illustrates the increasing displacement from the initial FePc structure. The limited number of evaluated structures does not support attribution of the dissolution-energy improvements to individual substituents or a causal structure-property relationship.

\section{Additional ablations and diagnostics}
\label{app:ablations}

\subsection{Proposal mechanism $\times$ adaptive region control}
\label{app:region_ablation}

Section~\ref{sec:region_ablation} compares two proposal mechanisms with and without the coupled TRACE region controller. The crossed design contains four configurations:
\begin{enumerate}
    \item TRACE proposer + adaptive TRACE region,
    \item TRACE proposer without adaptive region control,
    \item Graph-GA proposer + adaptive TRACE region,
    \item Graph-GA proposer without adaptive region control.
\end{enumerate}

For each objective and seed, the four configurations use the same $20$ evaluated initialization molecules. Active optimization then consists of $100$ iterations with batch size $q=10$, giving $20+100\times10=1{,}020$ cumulative oracle evaluations. We use five seeds on osimertinib MPO and scaffold hopping. Apart from the different batch size and number of active iterations, the region-enabled runs use the same controller settings as the primary benchmark in Section~\ref{sec:low_budget_benchmarks} (see Appendix~\ref{app:benchmark_implementation}). The molecular constraints, GP surrogate, and qLogEI acquisition procedure are also held fixed across the four configurations.

In a region-enabled configuration, the corresponding proposer supplies the finite proposal set $\mathcal{P}_t$, and admission follows Section~\ref{sec:trace_design}. In particular,
\[
\mathcal{S}_t=
\left\{
x\in\mathcal{P}_t\setminus\mathcal{X}(\mathcal{D}_t):
T(x,c_t)\geq\tau_t
\right\},
\qquad
|\mathcal{A}_t|=\min(k_t,|\mathcal{S}_t|),
\]
with $\mathcal{A}_t$ containing the $k_t$ molecules in $\mathcal{S}_t$ most similar to the current search center $c_t$ whenever $|\mathcal{S}_t|>k_t$. The cardinality budget $k_t$ adapts from the controller success/failure signal and $\tau_t$ follows from the coupling in Eq.~\eqref{eq:coupling}. For the TRACE proposer, the requested strict local proposal-pool size also follows the method definition, $\max(50,k_t)$.

In the region-disabled configurations, the corresponding proposal mechanism itself is unchanged, but the coupled adaptive-region control is removed: $k$ is held at $k_0$, no Tanimoto-similarity threshold is imposed, and no top-$k$ admission truncation is applied before the surrogate/acquisition stage. For the TRACE proposer, fixing $k=k_0$ also fixes the requested strict local proposal-pool size at $\max(50,k_0)$ rather than adapting it during the run. Other proposer-specific operations, molecular constraints, surrogate fitting, acquisition, and batch evaluation are unchanged. Thus, holding the proposer fixed refers to using the same proposal mechanism within each paired comparison; the candidate sets need not be identical because adapting candidate-search breadth is itself part of the region intervention.

The trajectories in Figure~\ref{fig:region_ablation} give the primary ablation result. With the TRACE proposer, adaptive region control improves the mean best-so-far trajectory on both objectives. The same intervention does not provide a corresponding benefit with Graph-GA proposals and produces mildly lower mean trajectories in the two cases studied. We next examine the internal controller behavior in the two region-enabled configurations to understand whether their optimization histories exhibit a corresponding difference.

\begin{figure*}[t]
    \centering
    \includegraphics[width=0.95\linewidth]{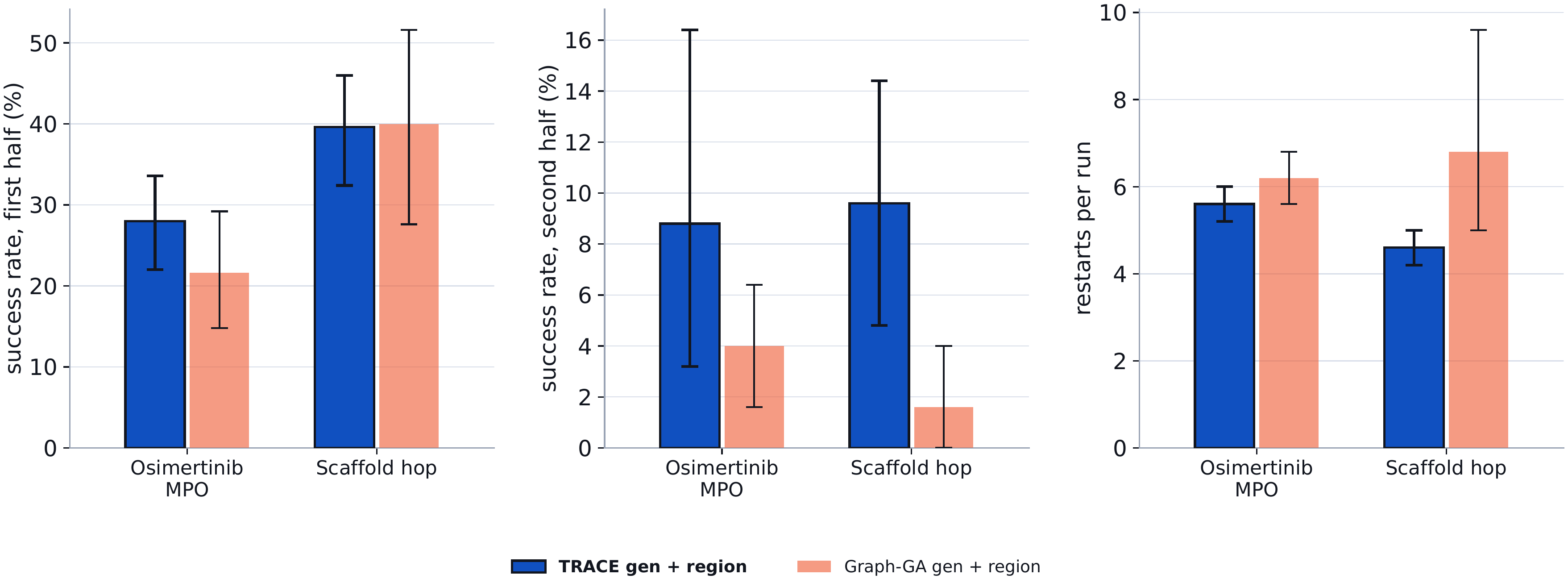}
    \caption{\textbf{Region controller diagnostics under the two proposal mechanisms.}
    Both configurations shown here enable adaptive $(\tau_t,k_t)$ control and differ in the proposal mechanism supplying $\mathcal{P}_t$. An iteration is counted as a controller success when
    $f(x_{t+1}^{\mathrm{best}})>f(x_t^{\mathrm{best}})+\varepsilon\max(1,|f(x_t^{\mathrm{best}})|)$
    with $\varepsilon=10^{-3}$, matching the success criterion that drives region adaptation. The first half comprises active iterations $1$-$50$, corresponding to the first $500$ active evaluations and ending at cumulative oracle call $520$; the second half comprises active iterations $51$-$100$, corresponding to the remaining $500$ active evaluations and ending at cumulative call $1{,}020$. A restart is counted whenever TRACE executes either restart pathway described in Section~\ref{sec:trace_design}: failure-driven contraction below the feasible batch floor or persistent candidate underfill after relaxed fallback generation. A restart resets $k$ to $k_0$ and may change the search center $c_t$ without discarding the global incumbent $x_t^{\mathrm{best}}$. Bars are means over five seeds and error bars are $95\%$ percentile-bootstrap intervals.}
    \label{fig:region_controller_diagnostics}
\end{figure*}

\begin{table}[t]
    \centering
    \caption{\textbf{Region controller diagnostics under the two proposal mechanisms.}
    Success rates and restart counts are measured in the two region-enabled configurations. Values are means over five seeds with descriptive $95\%$ percentile-bootstrap intervals in brackets.}
    \label{tab:region_controller_diagnostics}
    \small
    \begin{tabular}{llccc}
        \toprule
        & & \multicolumn{2}{c}{Region success rate (\%)} & Restarts\\
        \cmidrule(lr){3-4}
        Objective & Proposer & First half & Second half & per run\\
        \midrule
        \multirow{2}{*}{Osimertinib MPO}
          & TRACE & $28.0$ $[22.0,33.6]$ & $8.8$ $[3.2,16.0]$ & $5.6$ $[5.2,6.0]$\\
          & Graph-GA & $21.6$ $[14.8,28.4]$ & $4.0$ $[1.6,6.4]$ & $6.2$ $[5.6,6.8]$\\
        \midrule
        \multirow{2}{*}{Scaffold hopping}
          & TRACE & $39.6$ $[32.0,46.0]$ & $9.6$ $[4.8,14.4]$ & $4.6$ $[4.2,5.0]$\\
          & Graph-GA & $40.0$ $[27.6,51.6]$ & $1.6$ $[0.0,4.0]$ & $6.8$ $[5.0,9.6]$\\
        \bottomrule
    \end{tabular}
\end{table}

Figure~\ref{fig:region_controller_diagnostics} and Table~\ref{tab:region_controller_diagnostics} show how the controller behaves when paired with the two proposal mechanisms. During the first $50$ active iterations, there is no clear separation at the resolution of these five-seed summaries. On osimertinib MPO, the mean success rates are $28.0\%$ for the TRACE proposer and $21.6\%$ for Graph-GA; on scaffold hopping they are $39.6\%$ and $40.0\%$, respectively. The corresponding bootstrap intervals overlap in both cases.

The pattern differs in the second half of optimization, when improvements are less frequent for both proposers. The TRACE proposer retains mean controller-success rates of $8.8\%$ on osimertinib MPO and $9.6\%$ on scaffold hopping, compared with $4.0\%$ and $1.6\%$ for Graph-GA. Graph-GA also produces more restarts on average on both objectives, with the larger difference on scaffold hopping ($6.8$ versus $4.6$ restarts per run). Given only five seeds, these bootstrap summaries are best interpreted descriptively rather than as formal tests of pairwise differences.

Together, the later-run success rate and restart patterns are consistent with the interpretation in Section~\ref{sec:region_ablation}: the same adaptive controller interacts differently with the candidate distributions supplied by the two proposers. The TRACE proposer supplies structural variation around the current search center $c_t$, while the controller determines how much of that variation is admitted to BO through $\tau_t$ and $k_t$. A more dispersed proposal distribution can interact differently with this localization, which is consistent with the lower late-stage success rates and more frequent restarts observed for the Graph-GA pairing here. These diagnostics do not directly measure candidate density within the admissible region or establish a general causal mechanism, so we restrict the conclusion to the two objectives and five seeds studied.

\subsection{Batch size sensitivity on scaffold hopping}
\label{app:batch_sensitivity}

\begin{figure}[H]
    \centering
    \includegraphics[width=0.92\linewidth]{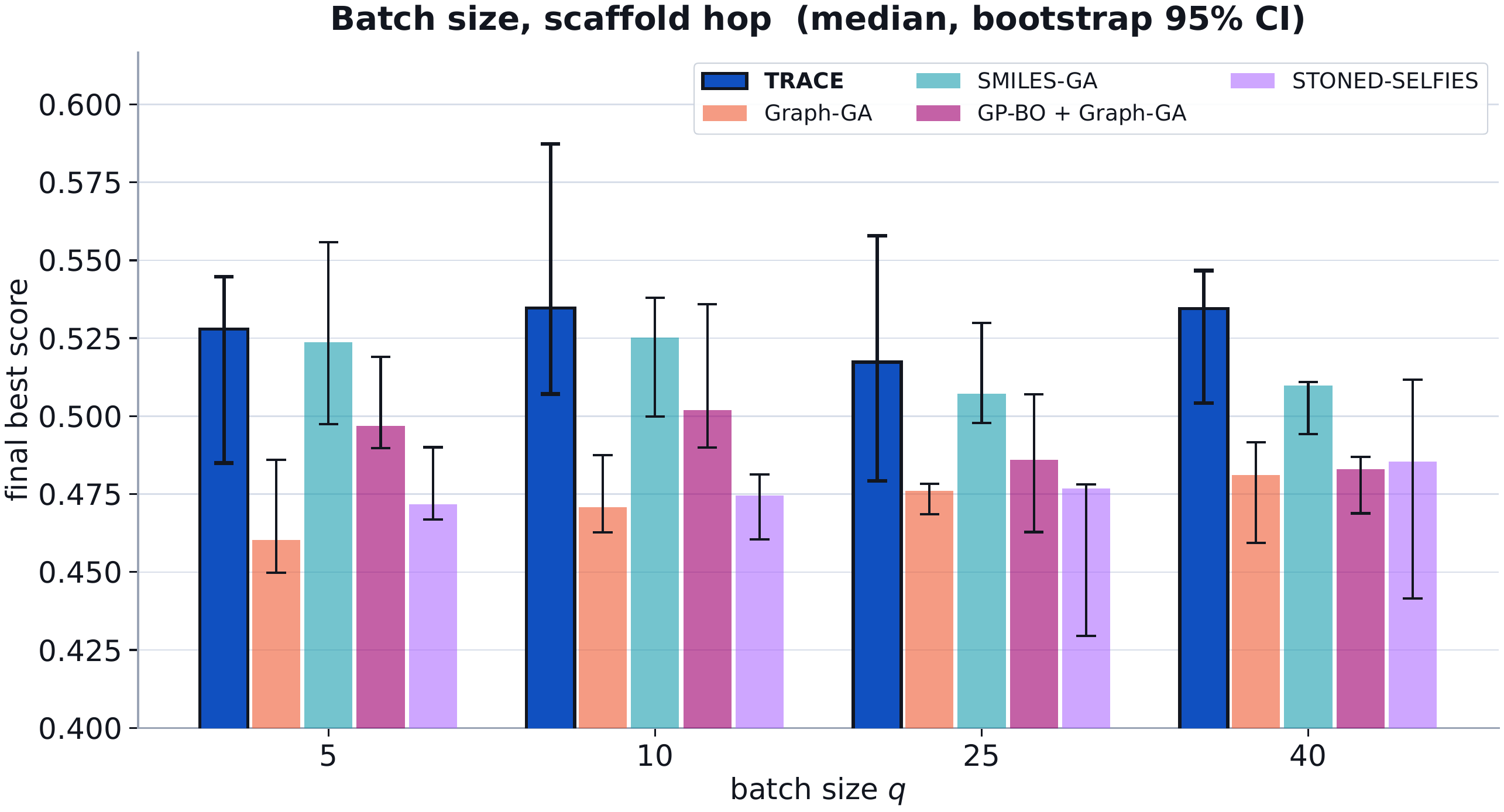}
    \caption{
    \textbf{Batch size sensitivity on scaffold hopping.}
    Final best score after 20 matched initialization evaluations followed by $1{,}000$ active oracle evaluations ($1{,}020$ cumulative evaluations) for batch sizes $q\in\{5,10,25,40\}$. Bars denote the median across five runs and error bars denote bootstrap $95\%$ intervals. TRACE has the highest observed median at each tested batch size. The modest variation across $q$ and substantial overlap of the uncertainty intervals indicate that the experiment mainly supports robustness across the tested settings, while differences among individual batch sizes are difficult to distinguish with five runs.}
    \label{fig:batch_sweep}
\end{figure}

TRACE updates its adaptive search region after each evaluated batch, so changing the batch size $q$ affects both the number of oracle evaluations between adaptive updates and the minimum region size needed to supply a complete batch, i.e., $\max(k_{\min},q)$. We examine sensitivity to these effects on the scaffold hopping objective using $q\in\{5,10,25,40\}$. For each seed, all methods start from the same 20 evaluated molecules and then receive exactly $1{,}000$ active unique-molecule oracle evaluations. This corresponds to 200, 100, 40, and 25 complete active batches for $q=5,10,25,40$, respectively, giving $1{,}020$ cumulative oracle evaluations including initialization in every case. We use five seeds and compare TRACE with Graph-GA, SMILES-GA, GP-BO + Graph-GA, and STONED-SELFIES at each batch size. Figure~\ref{fig:batch_sweep} reports the median final best objective value across runs together with bootstrap $95\%$ intervals.

TRACE's median final score varies only modestly across the sweep, from approximately $0.52$ to $0.54$, and is the highest observed median among the evaluated methods at each tested value of $q$ (Figure~\ref{fig:batch_sweep}). There is no clear monotone relationship between batch size and terminal performance. The median decreases slightly at $q=25$ and increases again at $q=40$, with overlapping bootstrap intervals across the TRACE settings. Overall, performance remains similar as the batch size varies from five to forty despite the corresponding changes in adaptive update frequency and feasible batch requirement.

Given that this sweep covers one objective and five runs per setting, its main value is as a robustness check for scaffold hopping over the tested range of $q$. Determining whether a particular batch size is generally preferable would require a broader study across objectives and additional runs.

\subsection{Penalized logP as a diagnostic objective}
\label{app:plogp}

We also evaluate penalized logP (pLogP), a long-standing molecular optimization benchmark that combines the octanol-water partition coefficient with synthetic-accessibility and large ring penalties \citep{muthyala2026generative}. The objective is known to favor increases in molecular size and related compositional changes, so high scores need not reflect broader molecular design quality \citep{gao2022sample}. To limit this size-seeking behavior, we impose the same 60-heavy-atom maximum on every method. We otherwise follow the primary benchmark schedule, using the same 20 evaluated initialization molecules for each seed followed by 20 active iterations with batch size $q=49$. Initialization is included in the budget, giving exactly 1,000 unique-molecule oracle evaluations per run, and we use five seeds.

\begin{figure}[t]
\centering
\includegraphics[width=0.9\linewidth]{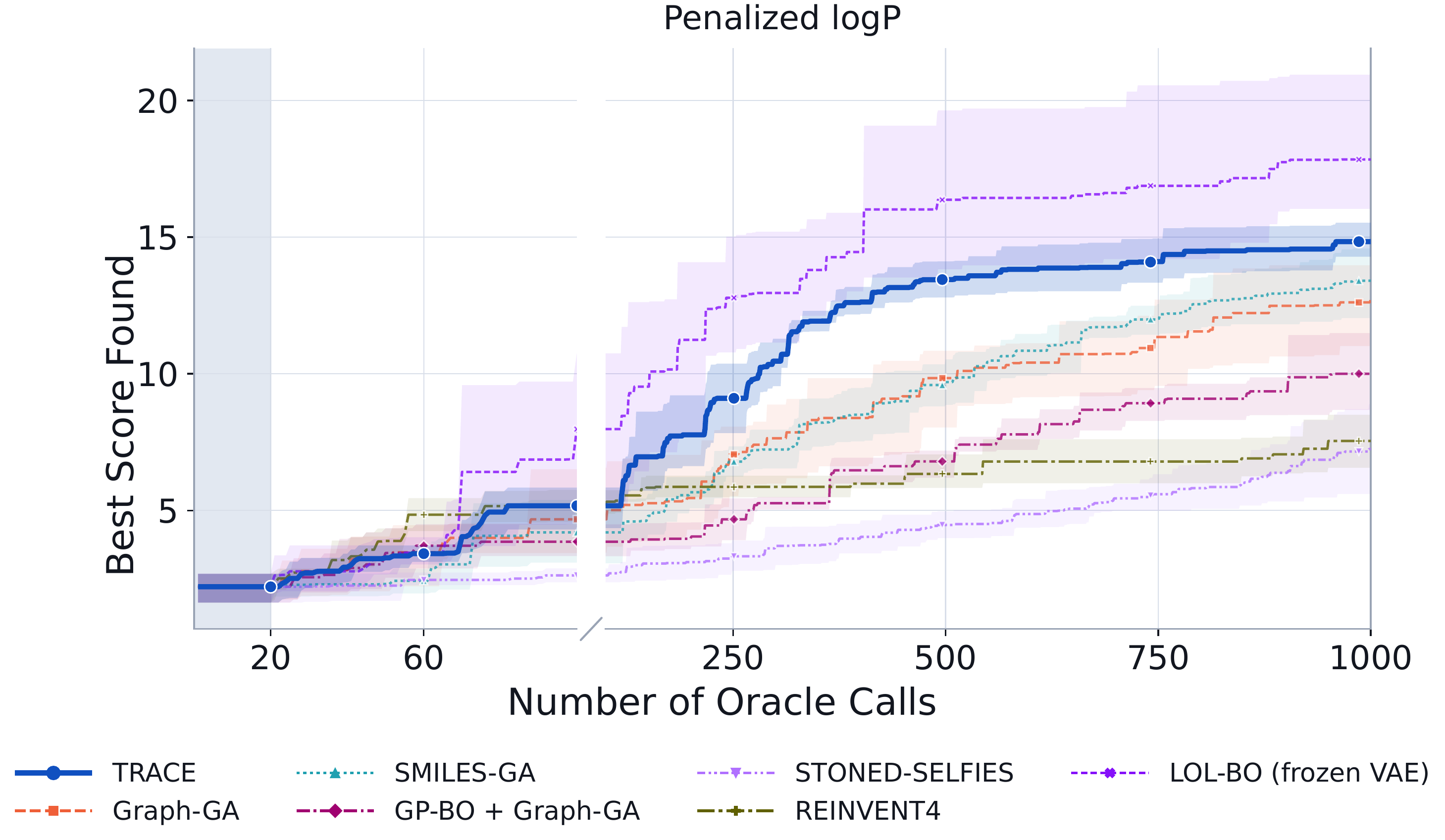}
\caption{\textbf{Penalized logP diagnostic.} Mean best-so-far pLogP score against the number of cumulative unique oracle evaluations under the matched 1,000-call budget. The common 20-molecule initialization is included in the total budget and shown by the gray region, after which optimization proceeds for 20 batches of $q=49$. The broken x-axis visually expands the early low-budget regime. Shaded regions denote 95\% percentile-bootstrap intervals over whole runs across five seeds.}
\label{fig:plogp_diagnostic}
\end{figure}

Figure~\ref{fig:plogp_diagnostic} shows a clear difference in the scores reached by the two leading methods. LOL-BO (frozen VAE) attains the highest observed pLogP performance, with a mean final best score of approximately $18.23$, compared with $14.84$ for TRACE. Inspection of the saved highest-scoring molecules shows extended hydrophobic chains among the LOL-BO (frozen VAE) solutions and, in some runs, unusual phosphorus-rich motifs. The final-best molecules from both methods also lie close to the imposed size limit, with mean heavy-atom counts of approximately $59.2$ for LOL-BO (frozen VAE) and $59.8$ for TRACE. The same tendency is visible across the full evaluated sets, where pLogP is positively associated with heavy-atom count for both methods (Spearman $\rho\approx0.63$ for LOL-BO (frozen VAE) and $\rho\approx0.84$ for TRACE).

Taken together, these diagnostics show that both methods exploit a size-dependent direction in pLogP, even with the 60-heavy-atom cap, while LOL-BO (frozen VAE) reaches higher scores under its search space. We do not interpret this difference as a general advantage of latent-space BO or a general limitation of TRACE. Rather, pLogP is useful here as a diagnostic because it makes visible how an objective with an easily exploited direction can interact with the chemistry accessible to a molecular generator. We therefore report it separately from the six-objective headline aggregate rather than use it as a measure of general molecular design performance.



\end{document}